\documentclass[lettersize,journal]{IEEEtran}
\usepackage{amsfonts,amssymb,amsmath}
\usepackage{bbm}
\DeclareMathOperator*{\argmax}{arg\;max} 
\usepackage{mathtools}
\usepackage{algorithm}
\usepackage[noend]{algpseudocode}

\usepackage{array}
\usepackage[caption=false]{subfig}
\usepackage{textcomp}
\usepackage{stfloats}
\usepackage{url}
\usepackage{verbatim}
\usepackage{graphicx}
\usepackage{threeparttable}
\usepackage{cite}
\usepackage{hyperref}
\usepackage{adjustbox}
\usepackage{siunitx}
\usepackage[capitalise]{cleveref}
\usepackage{multirow}
\usepackage{color,soul}
\begin{document}

\title{\huge Decision Making for Autonomous Driving in Interactive Merge Scenarios via Learning-based Prediction}

\author{Salar Arbabi, Davide Tavernini, Saber Fallah and Richard Bowden
	\thanks{Salar Arbabi, Davide Tavernini and Saber Fallah are with the Centre for Automotive Engineering, University of Surrey, Guildford, GU2 7XH, U.K. (e-mail: \{s.arbabi, d.tavernini, s.fallah\}@surrey.ac.uk).}
	\thanks{Richard Bowden is with the Centre for Vision, Speech and Signal Processing, University of Surrey, Guildford GU2 7XH, U.K. (e-mail: r.bowden@surrey.ac.uk).}
} 



\maketitle
\begin{abstract}

Autonomous agents that drive on roads shared with human drivers must reason about the nuanced interactions among traffic participants. This poses a highly challenging decision making problem since human behavior is influenced by a multitude of factors (e.g., human intentions and emotions) that are hard to model. 
This paper presents a decision making approach for autonomous driving, focusing on the complex task of merging into moving traffic where uncertainty emanates from the behavior of other drivers and imperfect sensor measurements. 
We frame the problem as a partially observable Markov decision process (POMDP) and solve it online with Monte Carlo tree search. The solution to the POMDP is a policy that performs high-level driving maneuvers, such as giving way to an approaching car, 
keeping a safe distance from the vehicle in front or merging into traffic. 
Our method leverages a model learned from data to predict the future states of traffic while explicitly accounting for interactions among the surrounding agents. 
From these predictions, the autonomous vehicle can anticipate the future consequences of its actions on the environment and optimize its trajectory accordingly. We thoroughly test our approach in simulation, showing that the autonomous vehicle can adapt its behavior to different situations. We also compare against other methods, demonstrating an improvement with respect to the considered performance metrics.
 
\end{abstract}

\begin{IEEEkeywords}
Autonomous driving, decision
making, Monte Carlo tree search, POMDP.
\end{IEEEkeywords}

\section{Introduction}
 
\IEEEPARstart{D}{riverless} vehicle technology has the potential to bring many social benefits, such as enhanced traffic safety and increased productivity \cite{fagnant2015preparing}. Modern autonomous vehicles are able to sense their local environment, recognize relevant objects, and make driving decisions that obey traffic rules \cite{ardelt2012highly, ziegler2014making}. Nevertheless, many situations encountered during daily driving continue to be challenging for autonomous vehicles, holding back their commercial and large-scale deployment. In particular, safely navigating traffic scenarios that involve interactions with human drivers requires the design of decision making algorithms that can reason about the uncertain motion of other vehicles while relying on noisy and incomplete sensor measurements.

Early-stage autonomous vehicles reacted to traffic situations based on a set of deterministic rules and heuristics \cite{montemerlo2008junior, kammel2008team, baker2008traffic}. 
These methods often have difficulty scaling up to complex interactive scenarios as explicitly programming a decision making agent to deal with changing environments and the diversity of human behavior is notoriously difficult, if not infeasible---it would require exhaustively
enumerating all the actions the agent can take in response to the situations that might arise (for discrete state/action spaces).  
Furthermore, in many scenarios, it is particularly important to take into account the future consequences of any action. For example, when performing a merge maneuver, the agent has to accelerate to match the speed of the oncoming traffic, adjust its speed for better positioning, and gradually move toward the main road. The risks and benefits associated with the maneuver, which may take several seconds to complete, can only be determined by anticipating how the traffic environment is likely to evolve over time. To this end, the agent relies on computational models that are capable of capturing complex and uncertain traffic dynamics for future prediction. 
These challenges make decision making a suitable candidate for the application of methods from machine learning and artificial intelligence that can produce probabilistic predictions and enable intelligent behavior to emerge automatically rather than be human-programmed for a number of predefined situations.

\begin{figure}
    \centering
    \includegraphics[trim={0cm 15.5cm 21.5cm 0cm}, clip=true,
    width=\linewidth]{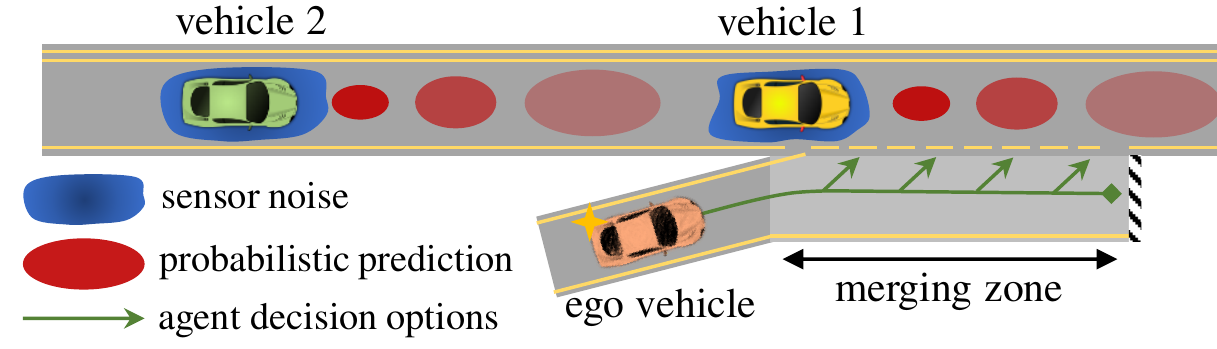}
    \caption{Example of a merging scenario with the ego vehicle (orange) approaching the merging zone. The vehicles on the main road can either yield to or ignore the ego vehicle.}\label{fig:agent_merge_example}
\end{figure} 

The central contribution of this study is a decision making algorithm for autonomous driving. We develop our approach around a ramp merging scenario (see \cref{fig:agent_merge_example}) in which there can be a high degree of uncertainty emanating from the behavior of other drivers and imperfect sensor measurements. The algorithm determines the autonomous vehicle's driving policies: closed-loop controllers that perform various high-level driving behaviors, such as keeping a safe distance from the car in front, performing a merge maneuver, or giving way to oncoming traffic. 
We formulate the decision making task as a partially observable Markov decision process (POMDP), which provides a theoretically-grounded framework for handling uncertainty. At each decision step, the POMDP problem is solved to determine the policy that optimizes the agent's trajectory with respect to safety and other relevant driving objectives.
Our solution method involves online planning with a Monte Carlo tree search (MCTS) whereby the potential consequences of the agent's actions are accounted for using an interaction-aware, probabilistic model of traffic dynamics.
The model is from our previous publication \cite{arbabi2022learning}, which combines neural networks with domain knowledge to generate accurate long-term predictions.

Specifically, our contributions are five-fold:
\begin{enumerate}
    \item We present a planning algorithm based on the POMDP framework, accounting for the uncertainty that originates from the behavior of other drivers and noisy onboard sensors (\cref{sec:pomdp_formulation}).
    \item We incorporate two heuristics in the agent's objective function to ensure driving safety (\cref{sec:reward_function}).
    \item We combine a learned model with the planner to consider the future outcomes of potential actions which can result from interactions among multiple drivers and the autonomous vehicle (\cref{sec:behavior_model}). 
    \item Through simulation-based case studies, we show that our approach generates appropriate driving behaviors to navigate various traffic situations (\cref{sec:qualitative}).
    \item We compare the proposed approach to several existing baselines, testing all methods in our traffic simulator to facilitate comparison (\cref{sec:quantitative}). 
\end{enumerate}
 
\section{Related Work}

There have been numerous works on decision making for autonomous driving, where approaches can be broadly distinguished by the assumptions made about the dynamics of the traffic environment and the degree to which uncertainties are considered.
Some methods do not account for the future motion of neighboring vehicles or assume that vehicles maintain a constant velocity and heading. Among the most common functional architectures employed are systems based on the Finite/Hybrid State Machine, whereby driving is divided into multiple discrete states, each initiating a distinct driving behavior such as lane keeping or intersection handling \cite{ziegler2014making, cosgun2017towards, noh2017decision}. Another approach is to emulate human drivers' preferences and decision making abilities with machine learning \cite{arbabi2020lane}. Alternatively, some methods involve predicting vehicle trajectories first and then using these predictions to plan collision-free paths for the autonomous vehicle~\cite{ardelt2012highly, bahram2014prediction}. These approaches can generate easy-to-interpret driving behaviors and benefit from a modular design, enabling rapid prototyping and development. However, they often have difficulty scaling up to scenarios where it is necessary to consider the coupled interactions among agents. 
 
Algorithms based on the framework of Reinforcement Learning (RL) have led to significant achievements in games \cite{mnih2015human, hausknecht2015deep, silver2017mastering} and are also being used to construct autonomous driving agents \cite{yildirim2022prediction, kuutti2019end, bouton2019cooperation}. 
In a typical RL setting, the agent interacts with the environment and learns a policy from past failures and successes.
Currently, the application of RL to autonomous driving requires that data generation is inexpensive and that there is no uncertainty about the state of the environment.  
However, collecting real-world driving data is often costly (and potentially dangerous). Moreover, as autonomous vehicles rely on noisy sensor measurements, critical information about the state can go uncaptured, including information about the intentions of other drivers.

An agent navigating a partially observable environment with stochastic dynamics can be modeled as a POMDP. In the POMDP framework, the agent has uncertain knowledge about the state of the environment, which is often represented as a probability distribution referred to as a belief state \cite{kaelbling1998planning}.
The efficacy of several decision making algorithms based on the POMDP formulation has been demonstrated through both simulation-based~\cite{sunberg2020improving} and real-world driving experiments~\cite{Galceran2017}.
Bouton et al. proposed a POMDP planner for navigating intersections \cite{bouton2017belief}, where the behavior of other drivers is considered to be the partially observable feature of the environment. A vehicle is assumed to move according to either a constant velocity model or a constant acceleration model, and the agent's belief is maintained using an Interacting Multiple Model (IMM). 
Another approach models interactions among the autonomous vehicle and other drivers via heuristic rules \cite{hubmann2018automated}. While the explicit consideration of uncertainty in these studies improves system robustness, one of their shortcomings is their reliance on hand-engineered (observation and transition) models, which may lack sufficient capacity to model nuanced social interactions on the road.

A more sophisticated motion model based on a Dynamic Bayesian Network that can capture interaction dynamics was used in \cite{gonzalez2019human}. Similar to our work, the model was used in a generative fashion to sample likely future observations. The agent's belief was represented by a Gaussian distribution and updated using an Extended Kalman Filter (EKF). 
While EKFs are commonly used in robotics, their requirement for prior knowledge of the transition and observation functions limits their capacity to model complex phenomena like driver behavior. In this work, we instead use non-linear transformations parameterized by recurrent neural networks (RNNs), which are known to be highly expressive models. 
Another approach is to perform planning by combining a neural-network policy and value function with the MCTS algorithm \cite{hoel2019combining}. Simulation experiments demonstrate that the proposed algorithm outperforms a vanilla MCTS baseline; however, the approach only considers the most likely state of the environment to inform agent actions. In contrast, our method uses sampled belief states to account for the agent's uncertainty about the state. 
 
\section{THEORETICAL BACKGROUND}

A Markov decision process (MDP) is a general formulation of sequential decision making under uncertainty. 
In an MDP, the agent has full knowledge of the environment's current state $s_t$, although the system can have stochastic dynamics. POMDPs extend MDPs to environments where the agent might not have direct access to the true state. Therefore, the current observation alone is insufficient for choosing optimal actions. Instead, the agent has to extract further information from the history of past actions and observations $h\doteq~\{a_0, o_1, ..., a_{t-1}, o_t\}$. 
To not have to keep track of arbitrarily long histories, it is often more straightforward to form and maintain a belief state $b \in \mathcal{B}$ based on the agent's experience thus far. Formally, a belief state is a probability distribution over state that captures the unobserved and uncertain aspects of the environment. A POMDP is generally defined by the tuple $\mathcal{(S, A, T, R, O, Z, \gamma)}$, where
$\mathcal{S}$ is the state space,  
$\mathcal{A}$ is the action space, 
$\mathcal{T}$ is a transition function, 
$\mathcal{R}$ is a reward function, 
$\mathcal{O}$ is the observation space, 
$\mathcal{Z}$ is an observation function, and 
$\mathcal{\gamma}$ is a discount factor \cite{kochenderfer2015decision}.  
Solving a POMDP problem amounts to finding the policy $\pi: \mathcal{B} \to \mathcal{A}$ that maximizes the expected cumulative reward. The value function $V_{\pi}(b)$ is the expected return when the agent starts from a belief $b$ and follows a policy $\pi$. There is at least one optimal policy that achieves the optimal value function, 
\begin{equation}
\pi^* \coloneqq \argmax_{\pi} \left(E\left[\sum_{t=0}^{\infty}\gamma^t\mathcal{R}(s_t, \pi(b_t))|b_0, \pi \right]\right) 
\end{equation}

A major drawback of POMDP models is the high computational expense involved in solving them, which stems from the need to maintain a belief state and use it to derive the optimal policy. Rather than searching for the globally optimal policy, online planners obtain a local estimate for the optimal value function by constructing a search tree of belief states from the current belief. To update the agent’s belief state, most online planners use recursive \textit{Bayes filters} with different approximate belief representations
(e.g., a particle filter \cite{silver2010monte, sunberg2020improving} or a Kalman filter \cite{gonzalez2019human}). 

\section{PROPOSED APPROACH}

In the use case investigated (see \cref{fig:agent_merge_example}), the autonomous vehicle is tasked with safely merging into moving traffic.
In this section, we start by describing our formulation of the merging problem as a POMDP, followed by details about the predictive model and the planning algorithm. 

\subsection{POMDP Formulation} \label{sec:pomdp_formulation}
 
\subsubsection{State space}

The state of the environment $s=\{x_e, \{x_v, \xi_v\}_{v\in V|v\neq e}\}$ consists of the physical state of the ego vehicle and other vehicles, $x_e$ and $x_v$, respectively, where $V=\{v_e, v_1, v_2, ..., v_n\}$ and $n$ is the number of vehicles present in the traffic scene.
The state vector also contains each driver's internal state $\xi_v$, which is a set of variables that encode latent factors like driver aggressiveness and attentiveness; the term is discussed in further detail in \cref{sec:behavior_model}.
A vehicle's physical state is given by
\begin{equation}  
    x_v =\{\mathrm{x}_v, \mathrm{y}_v, \mathrm{\dot x}_v\}
\end{equation}
where $\mathrm{x}_v$ and $\mathrm{y}_v$ are the vehicle's longitudinal and lateral positions, respectively, and $\mathrm{\dot x}_v$ is the vehicle's longitudinal speed. For notational simplicity, we will drop the subscript $v$ when not reasoning about a specific vehicle.

\subsubsection{Action space}

We consider a small set of predefined, closed-loop policies $\Pi$. These policies are low-level controllers that perform various high-level behaviors, such as giving way to an approaching vehicle, lane keeping, or performing a merge.
The idea of incorporating closed-loop policies in MDPs was put into a reinforcement learning theoretical framework by Sutton et al. to reduce the computational burden of planning and learning in large problems \cite{sutton1999between}.
Here, each policy is described by a tuple $(I, \pi, \Lambda)$, where $\pi$ is one policy from the set of available policies $\Pi_a$, $I$ is the \textit{initiation condition} specifying if $\pi$ is available for execution in the current state, and $\Lambda$ is the \textit{terminal condition} indicating when the policy must be terminated.

Conventionally, actions in POMDPs last for a fixed time interval: a discrete action $a$ taken at time $t$ affects the state at time $t+1$. 
Instead, in our approach, each policy can last for a variable number of time steps.
This characteristic is beneficial for two reasons. 
First, since the search tree can only be expanded following the termination of a policy, trees with a greater depth can be constructed (given the same compute budget) compared to planners that start a new sub-tree after each time step.
Second, the use of temporally extended policies
enables vehicle control at two levels of abstraction: low-level control actions that are produced at a near real-time rate (\SI{10}{\hertz} in our case) to react to immediate traffic events, and high-level decisions that are computed less frequently.\footnote{The frequency with which the agent makes decisions is a design choice that reflects a trade-off between computational expense and the agent's reaction time; for autonomous driving, values in the range of $\SI{0.5}{\hertz}-\SI{2}{\hertz}$ are commonly chosen in the literature \cite{gonzalez2019human, sunberg2020improving, bouton2017belief, hubmann2018automated, hoel2019combining}.}

Given a chosen policy $\pi$, the agent's action $a_t$ is determined according to $\pi(o_t)$ where $o_t$ is the current observation and the action is a tuple of continuous longitudinal acceleration and lateral speed $\mathrm{\ddot x}$ and $\mathrm{\dot y}$, respectively. 
The longitudinal acceleration is determined by an adaptive cruise controller (ACC) with adjustable setpoints, and the lateral speed is either set to $0$ when lane keeping or to a fixed value of $\dot y_{\mathrm{merge}}$ when merging. The ACC is implemented following the method from previous work \cite{arbabi2022learning}, which yields a fast, deterministic mapping $\pi: o_t \times \psi \rightarrow a_t$, where $\psi \in [0, 1]$ assigns the ACC's aggressiveness level. Moreover, the ACC is designed to guarantee that the agent never crashes into the vehicle in front. The agent can decide to increase or decrease $\psi$ by $\psi^{i+1} \leftarrow \psi^{i} + \Delta \psi$ every $\delta_t$ seconds. Higher values of $\psi$ result in more aggressive driving tendencies, like driving faster and closer to the vehicle in front (tailgating). \cref{table:policy_initiation_termination_conditions} provides a summary of the initiation and termination conditions for the set of five policies the agent can choose from, where values for $\psi$ are capped between 1 and 0 corresponding to the most aggressive and timid driving styles, respectively.  
 
\begin{table}
    \centering
    \footnotesize
    \begin{threeparttable}
        \caption
        {Initial and Terminal conditions for the policies} 	
        \centering
        \label{table:policy_initiation_termination_conditions}
        \begin{tabular}{|>{\raggedright}p{0.2\linewidth}|>{\raggedright}p{0.27\linewidth}|>{\raggedright\arraybackslash}p{0.30\linewidth}|}
             \hline   
             \centering Policy ($\pi$) & \centering Initial condition ($I$) & \centering\arraybackslash Terminal condition ($\Lambda$)\\  
             \hline\hline
             ``merge-in'' & agent in merging zone & agent arrived at the main road \\ 
             \hline
             ``give-way'' 
             & agent in merging zone
             & approaching vehicle has passed  \\
             \hline
             ``decrease ACC setpoint'' & $\psi > 0$ & after $\delta_t$ \\
             \hline
             ``increase ACC setpoint'' & $\psi < 1$ & after $\delta_t$  \\
             \hline
             ``maintain'' & - & after $\delta_t$  \\
             \hline
        \end{tabular}
    \end{threeparttable}
\end{table}
 
\subsubsection{Transition function}

In our approach, the transition function is defined by 
a generative model, $G$ , which is composed of two components: 1) a physical model that dictates how a vehicle's action affects its state, and 2) a behavior model that estimates the probability distribution over other drivers' actions (agent's actions are determined by its policy). The behavior model is learned offline from a dataset of vehicle trajectories. For physical modeling, as in existing work \cite{hoel2019combining, sunberg2020improving}, we model each vehicle as a point-mass and assume constant longitudinal acceleration and lateral speed over a time step of $\Delta t$. This is a reasonable simplification as $\Delta t$ is relatively small ($\SI{0.1}{\second}$), and in this paper, we focus on the high-level behavior generation of the ego vehicle and not its low-level control. Successive longitudinal position, longitudinal velocity, and lateral position of a vehicle, $\mathrm{x}'_v$, $\mathrm{
\dot x}'_v$, and $\mathrm{
y}'_v$, are updated according to,
\begin{align}
& \mathrm{x}'_v = \mathrm{x}_v + \mathrm{\dot x}_v\Delta t+ \frac{1}{2} \mathrm{\ddot{x}}_v \Delta t^{2}  
\\
& \mathrm{
\dot x}'_v = \mathrm{\dot x}_v + \mathrm{\ddot{x}}_v\Delta t 
\\
& \mathrm{
y}'_v = \mathrm{y}_v + \mathrm{\dot y}_v\Delta t  
\end{align}
where $\mathrm{\ddot x}_v$ and $\mathrm{\dot y}_v$ are the longitudinal and lateral components of the vehicle's action, respectively. 
 
\subsubsection{Reward function} \label{sec:reward_function}

The agent's goal is to merge onto the main road in a reasonable time while maintaining the safety of itself and its neighboring vehicles. The crash-free ACC ensures that the agent will never cause a collision with the vehicle in front.
Nevertheless, reckless behaviors such as cutting in front of a fast-moving car can lead to the abrupt deceleration of other drivers on the road or potentially result in a collision if there is an insufficient time gap for merging. 
Thus, we define the following reward function, 
\begin{equation} \label{equ:reward_function}
    \begin{split}
        \mathcal{R}(s) = 
        \lambda_1 R_{\mathrm{goal}}(s) + 
        \lambda_2 R_{\mathrm{delay}}(s) \\
        + \lambda_3 R_{\mathrm{hard\_brakes}}(s) + 
        \lambda_4 R_{\mathrm{safe}}(s)
    \end{split}
\end{equation}
where the scalar lambda weights can be adjusted to balance between different objectives. The individual terms are, 
\begin{align}
& R_{\mathrm{goal}}(s) = 
        \begin{dcases}
        1 & \text{if merge completed} \\
        0 & \text{otherwise}
        \end{dcases} \\
& R_{\mathrm{hard\_brakes}}(s) = \underset{v \in V|v\neq e}{\max}(\mathbbm{1}(\mathrm{\ddot x}_v \leq b_{\mathrm{hard}})) \\
& R_{\mathrm{safe}}(s) = \mathbbm{1}(\mathrm{TTC}_{v} < \mathrm{TTC_{min}} \vee
\mathrm{TIV}_{v} < \mathrm{TIV_{min}})
\end{align}
where $\mathbbm{1}$ denotes an indicator function that evaluates to $1$ if its argument is true and $0$ otherwise. The reward term $R_{\mathrm{delay}}$ punishes the agent with a small penalty following every decision step until the agent completes its merge maneuver, i.e., the agent is penalized for choosing actions that delay progress toward successful completion of the merge.
Additionally, $R_{\mathrm{hard\_brakes}}(s)$ and $R_{\mathrm{safe}}(s)$ penalize the agent for causing rapid deceleration of neighboring cars and for entering states that violate safety, respectively. The time to collision (TTC) and inter-vehicular time (TIV)
between the ego vehicle and its rear neighboring car on the main road are used to identify unsafe states:
\begin{align}
& \mathrm{TTC}_v = \frac{\mathrm{x}_e - \mathrm{x}_v}{\mathrm{\dot x}_v - \mathrm{\dot x}_e}
\\
& \mathrm{TIV}_v = \frac{\mathrm{x}_e - \mathrm{x}_v}{\mathrm{\dot x}_v} 
\end{align}
where TTC with a rear vehicle on the main road is defined as the time it takes for the vehicle to collide with the agent if they both were to continue at their present speed and drive in the same lane.

It should be noted that the reward term $R_{\mathrm{safe}}(s)$ plays a critical role in reducing the chance of the agent entering dangerous traffic situations, such as recklessly pulling in front of another car. We have found that the learned behavior model fails to accurately predict the outcome of the agent's actions when operating in near-collision situations. The model performs poorly in such cases because the driving dataset $\mathcal{D}$ used for model training contains only nominal (i.e., collision-free) operating conditions.
In other words, while the agent's actions are deemed reasonable under the learned model, they can be catastrophic when executed in the environment---this is a drawback of planning using our model. Luckily, as we show later, discouraging the agent from entering near-collision situations via the reward term $R_{\mathrm{safe}}(s)$ is a simple and effective way of dealing with the imperfections of the learned model.  


\subsubsection{Observation space}

The agent's observation $o$ consists of the physical state of all the vehicles in the scene, i.e., $o=\{x_v\}_{v \in V}$.
The drivers' internal states are considered hidden from the agent but can be estimated from observations.
Consideration of occlusions, e.g., due to trees or other objects blocking the agent's view, is out of the scope of this paper. 

\subsubsection{Observation function}

The observation function $\mathcal{Z}(o|s, a, s')$ defines the probability of receiving an observation $o$ when entering a new state $s'$. Rather than specifying an explicit representation of the observation distribution \cite{bouton2017belief, sunberg2020improving}, we use 
the generative model $G$ to sample a successor observation and state $s_{t+1}, o_{t+1} \sim G(s_t, a_t)$. Further details are provided in the following sections.

\subsection{Behavior Model} \label{sec:behavior_model}
 
We model the behavior of vehicles on the main road using an extended version of the Intelligent Driver Model (IDM) \cite{treiber2000congested}. 
Given a set of driver parameters and the current traffic state, the IDM outputs driver accelerations over time that provide a trade-off between a driver's goal of reaching a desired speed and keeping a safe distance from the vehicle ahead. 
The set of driver parameters $\theta_{\mathrm{IDM}} = \{v_{\mathrm{des}}, d_{\mathrm{min}}, T_{\mathrm{des}}, a_{\mathrm{max}}, b_{\mathrm{max}}\}$ consists of the driver's desired speed $v_{\mathrm{des}}$, minimum separation distance $d_{\mathrm{min}}$, time gap $T_{\mathrm{des}}$, maximum acceleration $a_{\mathrm{max}}$ and maximum deceleration $b_{\mathrm{max}}$. 

A driver's longitudinal acceleration is determined by
\begin{equation}  \label{equ:idm_action}
\begin{aligned} 
\mathrm{\ddot x^t} = a_{\mathrm{max}}\left(1-\left(\frac{\mathrm{\dot x^t}}{v_{\mathrm{des}}}\right)^4-\left(\frac{d_{\mathrm{des}}(\mathrm{\dot x^t}, \mathrm{\Delta \dot x^t})}{d^t}\right)^2\right)
\end{aligned}
\end{equation}
where $\mathrm{\dot x^t}$ is the vehicle's speed, and $\mathrm{\Delta \dot x^t}$ and $d^t$ are the relative speed (approach rate) and distance headway to the vehicle ahead, respectively. The desired distance $d_{\mathrm{des}}$ is given by
\begin{equation} \label{equ:des_d}
\begin{aligned} 
d_{\mathrm{des}}\mathrm{(\dot x^t, \Delta \dot x^t)} = d_{\mathrm{min}} + T_{\mathrm{des}}\mathrm{\dot x^t} + \frac{\mathrm{\dot x^t\Delta \dot x^t}}{2\sqrt{a_{\mathrm{max}}b_{\mathrm{max}}}}
\end{aligned}
\end{equation}

We also introduce two additional parameters, $w_l$ and $w_m$, to model the degree of a driver's attentiveness toward its front neighbor and the autonomous car on the on-ramp, respectively. 
For a driver attending to the autonomous car, $w_m\approx1$, while $w_m$ approaches zero for a non-attentive driver. The overall acceleration of the vehicle is given by the sum of an acceleration determined when the vehicle attends to its front neighbor and when it attends to the agent, $\mathrm{\ddot x}= w_l \mathrm{\ddot x_l} + w_m \mathrm{\ddot x_m}$. 
We refer to the parameters $\xi=\{\theta_{\mathrm{IDM}}, w_l, w_m\}$ as the driver's internal state to emphasize that they cannot be measured directly. The goal is to estimate, for each vehicle, the values of these parameters from past observations. 
In the following section, we provide an overview of the model operations involved in estimating the parameters in $\xi$. We direct the reader to the original publication for the complete details \cite{arbabi2022learning}.   

\subsubsection{Approximate belief inference} \label{sec:belief_inference}

\begin{figure*}
    \centering
    \includegraphics[trim={0cm 11.7cm 12cm 0.cm}, clip=true,
    width=0.8\linewidth]{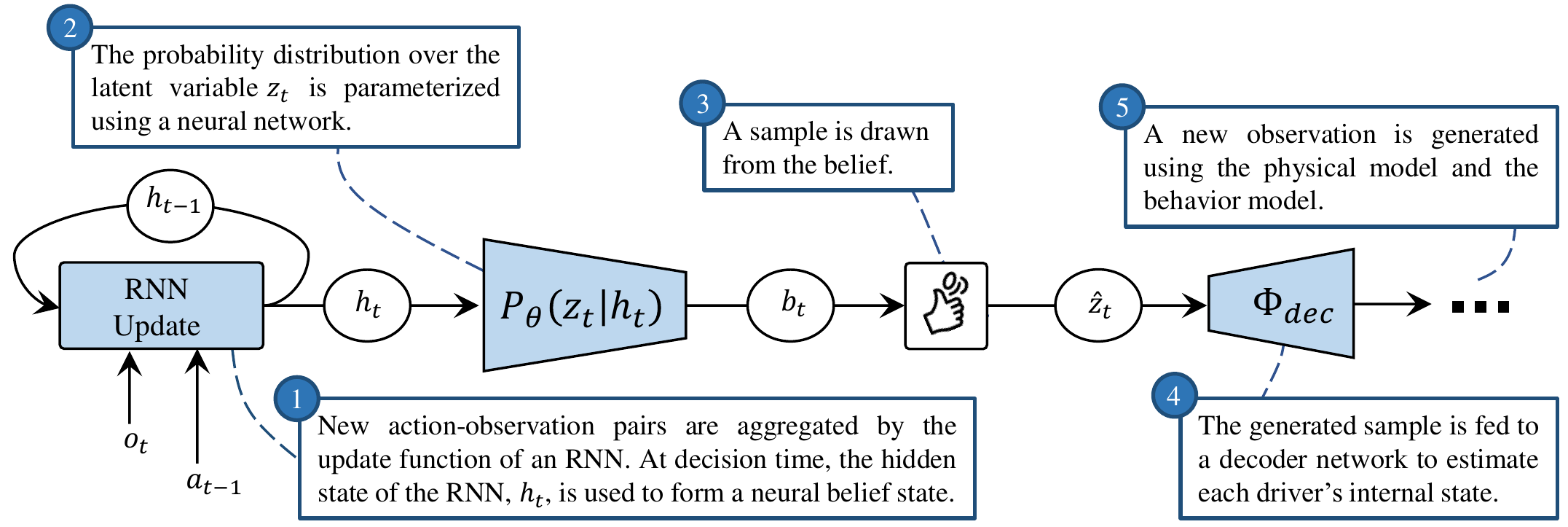}
    \caption{Flow diagram showing how a neural belief state is formed and used for prediction. Follow the panels for a brief explanation of our methodology. The model components involved (shown in blue) are neural networks whose parameters are learned from a dataset of vehicle trajectories collected from a traffic simulator. A description of the dataset and the method of data collection is detailed in \cite{arbabi2022learning}.}\label{fig:belief_module}
\end{figure*} 

We seek to infer belief states that can hold sufficient information for generating accurate, long-term predictions. 
A schematic diagram describing our method of forming compact belief representations for planning is shown in \cref{fig:belief_module}.
Let $z_t \in \mathbb{R}^{n_z}$ be a low-dimensional latent variable that captures unobserved and uncertain aspects of driver behavior.
With a slight abuse of notation, we use the learned neural network $P_{\theta}(z_t|h_t)$ to produce a surrogate representation for the belief and denote it as $b_t$, where $h_{t}$ is a vector summarizing the tracked action-observation history. Therefore, $b_t$ is not an explicit probability distribution over state $s_t$, but rather servers as an approximation that is inferred using a learned model. Samples drawn from the belief distribution can then be mapped to drivers' internal states via the decoder network $\Phi_{\mathrm{\textit{dec}}}$.  

More specifically, in this work, the probability distribution over $z_t$ is a diagonal Gaussian whose mean $\mu_t \in \mathbb{R}^{n_z}$ and variance $\sigma_t^2 \in \mathbb{R}^{n_z}$ are estimated by neural networks. We encode the history into a fixed length vector $h_{t}$ using the RNN variant Long Short-Term Memory (LSTM)~\cite{hochreiter1997long}.
We define a neural belief state as $b_t := \{\mu_{\mathrm{nn}}(h_{t, v}), \sigma_{\mathrm{nn}}(h_{t, v}), h_{t, v}\}_{v\in V|v\neq e}$, where for a vehicle $v$, $h_{t, v}$ is the conditional variable, and $\mu_{\mathrm{nn}}(h_{t, v})$ and $\sigma_{\mathrm{nn}}(h_{t, v})$ are neural network functions of $h_{t, v}$.
We include $h_{t, v}$ as part of the belief state so that there is a direct path from $h_{t, v}$ to the drivers' internal states; this is essentially granting the agent access to the memory, which reduces the chance of information bottlenecks forming as a result of compressing history into a low-dimensional latent variable $z_{t, v}$.

\subsubsection{Belief update}

At the start of a driving episode, when the agent has not received any observations, its belief is assumed to have a uniform distribution. Intuitively, a uniform distribution expresses the agent's lack of knowledge about other drivers prior to receiving any observations. 
As the episode proceeds, new action-observation pairs are integrated through the recurrent update function $h_t = \mathrm{RNN}(h_{t-1}, a_{t-1}, o_t)$. At a decision time $t$, the RNN's state $h_t$ is transformed to a probability distribution over $z_t$. The above procedure is performed for all the vehicles in the scene to form a neural belief state. 
Note that we do not perform belief inference with every incoming action-observation pair, as doing so would result in an unnecessary computational burden. Instead, $P_{\theta}(z_t|h_t)$ is only computed when the agent needs to make a decision and thus needs access to the current belief. 

\subsection{Planning Algorithm} \label{sec:planning_algorithm}

In the proposed decision making approach for autonomous driving, the agent’s decisions are determined by a POMDP-based planner, which evaluates a set of decision options and selects the one with the highest value for the agent to execute. The decisions of the agent are closed-loop policies, as discussed previously. The new approach will be referred to as Latent Variable Tree (LVT), which starting from a root node, constructs a search tree of policy and observation nodes by selecting policies that balance between exploration and exploitation. 
 
While the observations are generated by the model, the policies are selected by a particular bandit algorithm called UCB1 \cite{kocsis2006bandit} (UCB stands for Upper Confidence Bounds), which selects decisions according to:
\begin{equation}
\pi = \argmax_{\pi \in \Pi_a} V(b, \pi) + c\sqrt{\frac{logN(b)}{N(b, \pi)}}  \label{equ:ucb1}
\end{equation}
where $N(b, \pi)$ counts the number of times $\pi$ has been selected from belief $b$, $N(b) = \sum_{\pi \in \Pi_a} N(b, \pi)$ and 
$c > 0$ is a tunable parameter that dictates the amount of exploration in the tree. The UCB algorithm focuses the search on the probable and promising regions of the search space (e.g., regions that are unlikely to result in a dangerous traffic situation).

LVT shares the same structure as other online planners that combine UCB tree search with Monte Carlo simulations \cite{silver2010monte, sunberg2018online}, where each planning iteration consists of four stages: tree traversal, node expansion, rollout, and backpropagation. 
More iterations generally improve performance, although planning can be interrupted at any point to obtain an approximate solution. The fundamental distinction between LVT and existing planners is that LVT captures the uncertainty in the belief states using neural networks instead of particle ensembles. The way in which the beliefs are represented and updated is illustrated in \cref{fig:belief_module}. 
 
\subsubsection{Rollouts}

Rollouts provide a simple method for evaluating a state. To perform a single step of a rollout in our implementation, first, a policy is selected at random (i.e., uniform random selection) and a sample is drawn from the current belief. Next, the policy is applied to propagate the traffic state forward in time until the policy's termination condition is reached. Sequences of future states are generated by repeating this process, where each rollout starts from a newly expanded tree node and continues until a discount horizon or a termination condition is reached. We note that while policy selection is random, the future outcomes of actions are not.
We use the generative model $G$ to sample likely future traffic states. The model captures vehicle interactions (i.e., how the vehicles on the road react to the actions of one another), which is crucial for obtaining accurate value estimates.

\subsubsection{Planning in continuous spaces} \label{sec:DPW}

The vanilla MCTS and its extensions that address partial observability \cite{silver2010monte} cannot be effectively applied to problems with large or continuous spaces. 
For the latter, the probability of visiting the same state twice is infinitesimally small, resulting in a search tree that grows in width but does not grow beyond the first layer in depth; a wide, shallow tree cannot evaluate the long-term future consequences of actions, thus producing inaccurate value estimates. 
To accommodate the continuous state and observation spaces of our vehicle merging domain, we use a technique called double progressive widening (DPW) \cite{sunberg2018online}. This method enables the construction of deeper search trees while progressively \textit{unpruning} the lower parts of the tree to permit further exploration. 

\subsubsection{Description of planning algorithm}

\begin{figure}
    \centering
    \includegraphics[trim={0cm 8.5cm 21.5cm 0.cm}, clip=true,
    width=0.95\linewidth]{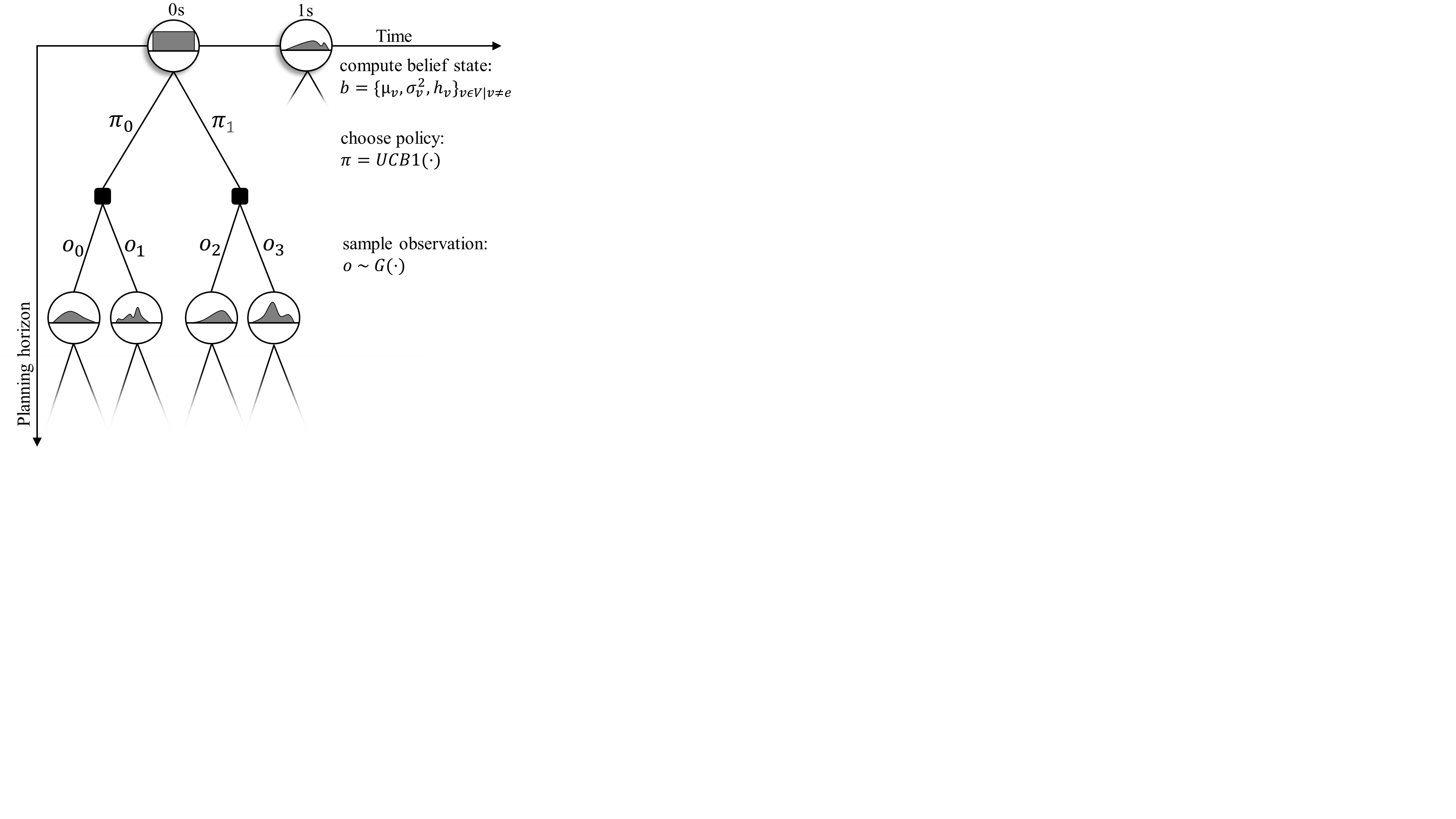}
    \caption{Example search tree for evaluating each available policy. A search tree is constructed using Monte Carlo simulations.}\label{fig:lvt_architecture}
\end{figure}   

\begin{algorithm} 
\caption{Latent Variable Tree}\label{alg:LVT} 
\begin{algorithmic}
\Procedure{Plan}{$\eta_0, b_0$}
    \For {$i \in 1:n$ }
        \State {\textsc{SIMULATE}$(\eta_0, b_0, d_{\mathrm{max}§})$}
    \EndFor
\State {$\pi^* \leftarrow \underset{\pi}{\argmax}\;V(\eta\pi)$} 
\State \Return {$\pi^*$} 
\EndProcedure
\vspace{0.2cm}
\Procedure{ActionProgWiden}{$\eta$}
    \If {$|C(\eta)| \leq k N(\eta)^{\alpha}$}
        \State {$\Pi_a \leftarrow $ \textsc{APPLICABLE}$(\eta)$}
        \State {$\pi \leftarrow \textsc{Next}(\Pi_a)$}
        \State {$C(\eta) \leftarrow C(\eta)\cup \{\pi\}$}
    \EndIf
    \State \Return {$\underset{\pi \in C(\eta)}{\argmax}\;V(\eta\pi) + c\sqrt{\frac{log\;N(\eta)}{N(\eta \pi)}}$} 
\EndProcedure
\end{algorithmic}
\end{algorithm}

An illustration of the planner is shown in \cref{fig:lvt_architecture}, and the complete algorithm is outlined in \cref{alg:LVT,,alg:simulate}. The following notations are used in the listings and text: $\eta = (\pi_0, o_1, ..., \pi_{t-1}, o_t)$ denotes a history of policies and observations, and $\eta \pi$ and $\eta \pi o$ are shorthand for sequences with $\pi$ and $(\pi, o)$ appended to the end; $\tau_{ao}$ is the snippet of the recent action-observation history; $d$ is the remaining tree depth to explore; $C$ is a list of a node's children; $N$ counts the number of node visits; $V$ is a value function estimate; and $B$ is the belief associated with a node. 

The \textsc{PLAN} procedure is called from the initial history $\eta_0$ and belief $b_0$; with each planning iteration, the tree is constructed incrementally (precisely one new node gets added to the tree per iteration). The first step in \textsc{SIMULATE} outputs a set of available candidate policies $\Pi_a$, where policy availability depends on the current traffic state. For example, if the agent has not yet reached the merging zone, a ``merge-in'' policy would not be applicable.
Next, a policy $\pi$ is chosen according to the function \textsc{ActionProgWiden}, and then depending on the node statistics and the observation widening parameters (line 5), either a new observation is generated or one is sampled from the set of existing observation child nodes $C(\eta\pi)$, where $k>0$ and $\alpha \in (0, 1)$ are parameters that control the rate of tree widening. 
To generate a new observation $o$, the generative model $G(\cdot)$ is applied repeatedly, propagating the traffic state forward until the policy $\pi$ is terminated. Next, the belief is updated by incorporating the history $\tau_{ao}$. Existing observations are sampled uniformly from $C(\eta\pi)$.
If the observation is newly generated, it is added to the tree as a leaf node (line 20), and a rollout is applied to estimate its value.
Otherwise, \textsc{SIMULATE} is called recursively until a leaf node is reached. Finally, the node statistics are updated. 

\begin{algorithm} 
\caption{Simulate Procedure}\label{alg:simulate}
\begin{algorithmic}[1]
\Procedure{Simulate}{$\eta, b, d$}
\If {$d=0$}
    \State \Return {$0$} 
\EndIf
\State {$\pi \leftarrow $ \textsc{ActionProgWiden}$(\eta)$}
\If {$|C(\eta\pi)|\leq k N(\eta\pi)^{\alpha}$}
    \State $s, o\leftarrow$ sample $s$ from belief $b$  
    \State $\tau_{ao} \leftarrow \emptyset$ 
    \Comment{$\text{empty list}$}
    \State $t \leftarrow 0$
    \While {$\pi$ is not terminated}
        \State $a_t \leftarrow \pi(o_t)$
        \State $s_{t+1}, o_{t+1} \sim G(s_t, a_t)$
        \State {\text{append }$(a_t, o_{t+1})$\text{ to }$\tau_{ao}$}
        \State $t \leftarrow t + 1$
    \EndWhile
     \State $s', o  \leftarrow s_{t}, o_{t}$
     \State $b' \leftarrow \textsc{RnnUpdate } (b, \tau_{ao})$ 
\Else
    \State $o \leftarrow \text{sample uniformly from }C(\eta\pi)$ 
    \State $b' \leftarrow B(\eta\pi o)$ \Comment{select a belief state}
    \State $s' \sim b'$  
\EndIf

\If {$o \notin C(\eta\pi)$}
    \State $C(\eta\pi) \leftarrow C(\eta\pi) \cup \{o\}$
    \State $r_{\mathrm{total}} \leftarrow \mathcal{R}(s') + \gamma \cdot \textsc{Rollout}(\eta\pi o,s',d-1)$
\Else
    \State $r_{\mathrm{total}} \leftarrow \mathcal{R}(s') + \gamma \cdot \textsc{SIMULATE}(\eta\pi o, b', d-1)$  
\EndIf 
\State $N(\eta) \leftarrow N(\eta) + 1$
\State $N(\eta\pi) \leftarrow N(\eta\pi) + 1$
\State $V(\eta\pi) \leftarrow V(\eta\pi) + \frac{r_{\mathrm{total}}-V(\eta\pi)}{N(\eta\pi)}$
\State \Return {$r_{\mathrm{total}}$} 
\EndProcedure
\end{algorithmic}
\end{algorithm}

\section{Experimental Setup}

This section provides a description of how the proposed approach is tested in a traffic simulator and how the baseline methods are implemented. The choice of evaluation baselines considered was inspired by Sunberg et al. \cite{sunberg2020improving}.

\subsection{Simulation Environment} \label{simulation-environment}

We evaluate our system via multi-agent traffic simulations. At the start of a driving episode, we populate the road with drivers whose aggressiveness ranges from the timidest to the most aggressive. Within the simulator, a driver's aggressiveness level $\psi \in [0, 1]$ is assigned either manually or randomly, depending on the experiment. As in \cite{bouton2017belief, hoel2019combining}, we apply a small perturbation to the position measurements of all vehicles (except for the ego car) to simulate sensor noise from the autonomous vehicle's perception. The noise applied to the longitudinal and lateral vehicle positions are $\epsilon_x\sim \mathcal{N}(0,\sigma_{x})$ and $\epsilon_y \sim \mathcal{N}(0, \sigma_{y})$, respectively.  
The longitudinal acceleration of vehicles is determined using the IDM \cite{treiber2000congested} and to simulate driver attentiveness, we adapt the Cooperative Intelligent Driver Model (C-IDM) proposed in \cite{bouton2019cooperation}. C-IDM uses estimates of time to merge $(\mathrm{TTM})$ for a vehicle on the main road $(\mathrm{TTM}_v)$ and the agent $(\mathrm{TTM}_e)$ to determine whether a main road driver intends to attend and yield to the agent on the on-ramp:
\begin{itemize}
    \item If  $\mathrm{TTM}_e < (1-\psi_v) \cdot \mathrm{TTM}_v$, the vehicle on the main road follows the IDM by considering the projection of the merging agent on the main road as its front neighbour, therefore exhibiting a yielding behaviour. 
    \item If $\mathrm{TTM}_e \geq (1-\psi_v) \cdot \mathrm{TTM}_v$, the vehicle on the main road follows the standard IDM, therefore exhibiting a passing behaviour. 
\end{itemize}
 
The method we use to evaluate the performance of the agent proceeds as follows. When an episode begins, the agent starts with a uniform distribution as its belief. From this belief, the agent chooses the best available policy and executes it in the environment. The environment's state evolves forward in time for the duration of a decision step following each vehicle's action commands. At the next decision step, the belief is updated with the latest observations and used by the agent to compute its next policy. In our simulator, observations are received by the agent every \SI{0.1}{\second}, actions are applied at a rate of \SI{10}{\hertz}, and a new policy is chosen immediately after the termination of the current one. The configuration parameters for the traffic simulator are listed in \cref{table:simulation_parameters}. 

\subsection{Baseline Methods} \label{sec:baseline_methods}
 
We compare our method with the following baselines:
\subsubsection{Omniscient}
We use this agent, which can foresee the future with perfect accuracy, to establish an approximate upper bound on performance. 

\subsubsection{MCTS}
This is the standard MCTS combined with progressive widening to deal with a continuous state space, as described in \cref{sec:DPW}.
Contrary to our approach, which uses a learned model to form beliefs about the internal state of other drivers, this agent is more conservative as it is unable to reduce uncertainty by learning about other drivers.
The implementation of the MCTS algorithm is from \cite{rl-agents}.

\subsubsection{MCTS-normal} 
This agent believes all drivers follow a \textit{normal} driving style (i.e., $\psi=0.5$). However, the simulated drivers in our experiments have different driving styles, from timid to aggressive. 

\subsubsection{QMDP} This method is based on the QMDP approximation \cite{littman1995learning}. 
While the QMDP implementation in this work still benefits from performing model-based rollouts, the approximation amounts to planning as if the uncertainty in the agent's current belief will vanish in subsequent steps. As a consequence, the QMDP agent has no incentive to take information-gathering actions that might be part of the optimal policy. In the merging scenario, for example, the agent can benefit from triggering reactions from other drivers by speeding up when approaching the merging zone; a normal driver is likely to yield to the agent, while an aggressive driver would remain unaffected. 

\subsubsection{Rule-based} This agent uses a simple, rule-based strategy to make decisions. 
The rules are based on the TTC and TIV metrics, two widely used safety heuristics in the automated driving domain \cite{bahram2014prediction, Evre2014, nilsson2013strategic}. Below is an explanation of the algorithm. If the agent has entered the merging zone, and TTC and TIV values exceed the thresholds $\mathrm{TTC_{safe}}$ and $\mathrm{TIV_{safe}}$, then the agent chooses the ``merge in'' policy. Otherwise, the agent gives way to the vehicles on the main road. In the case where the agent initially decides to merge, but later in the episode, the TTC and TIV values fall below their set thresholds, the agent switches back to the ``give-way'' policy. Regardless of which policy the agent chooses, its longitudinal acceleration is determined by an ACC with a fixed setpoint of $\psi=0.5$.
 
\begin{table}[t]
    \centering
    \begin{threeparttable}
        \caption
        {Various simulation and agent configuration parameters} 
        \label{table:simulation_parameters}
        \centering
        \begin{tabular}{p{3.7cm} p{1cm} p{2cm}}
            \hline
            Parameter & Symbol & Value\\  
            \hline
            Simulation step size & $\Delta t$ & \SI{0.1}{\second}\\
            Decision step size & $\delta_t$ & \SI{1}{\second}\\
            Lateral speed & $\dot y_{\mathrm{merge}}$ & \SI{0.75}{\meter\per\second}\\
            Longitudinal position noise & $\sigma_{x}$ & \SI{1}{\meter} \\
            Lateral position noise & $\sigma_{y}$ & \SI{0.2}{\meter} \\
            Main road length & - & \SI{500}{\meter}\\
            On-ramp length & - & \SI{200}{\meter}\\
            merging zone length & - & \SI{100}{\meter}\\
            Min time to collision & $\mathrm{TTC_{min}}$ & \SI{3.3}{\second}\\
            Min inter-vehicular time & $\mathrm{TIV_{min}}$ & \SI{1.3}{\second}\\
            Safe time to collision & $\mathrm{TTC_{safe}}$ & \SI{5.6}{\second}\\
            Safe inter-vehicular time & $\mathrm{TIV_{safe}}$ & \SI{2.5}{\second}\\
            Merge completion reward & $\lambda_1$ & $+5$\\  
            Delay penalty & $\lambda_2$ & $-0.3$\\  
            Hard braking penalty & $\lambda_3$ & $-5$\\ 
            Safety violation penalty & $\lambda_4$ & $-10$\\  
            Hard braking & $b_{\mathrm{hard}}$ & \SI{-4}{\meter\per\second\squared}\\
            \hline
             Discount factor & $\gamma$ & 0.9\\
             ACC step size & $\Delta \psi$ & 0.3\\
             Number of planning iterations & - & $1, 4, 16, ..., 1024$ \\
             Exploration param. & c & 5\\
             PW linear param. & $k$ & 2\\
             PW exponent param. & $\alpha$ & 0.1\\
             Max depth & $d_{\mathrm{max}}$ & 15 \\
            \hline
        \end{tabular}
    \end{threeparttable}
\end{table}

\begin{figure}[t]
    \centering
    \includegraphics[width=0.9\linewidth, trim={0.2cm 0.3cm 0cm 0.2cm}, clip=true]{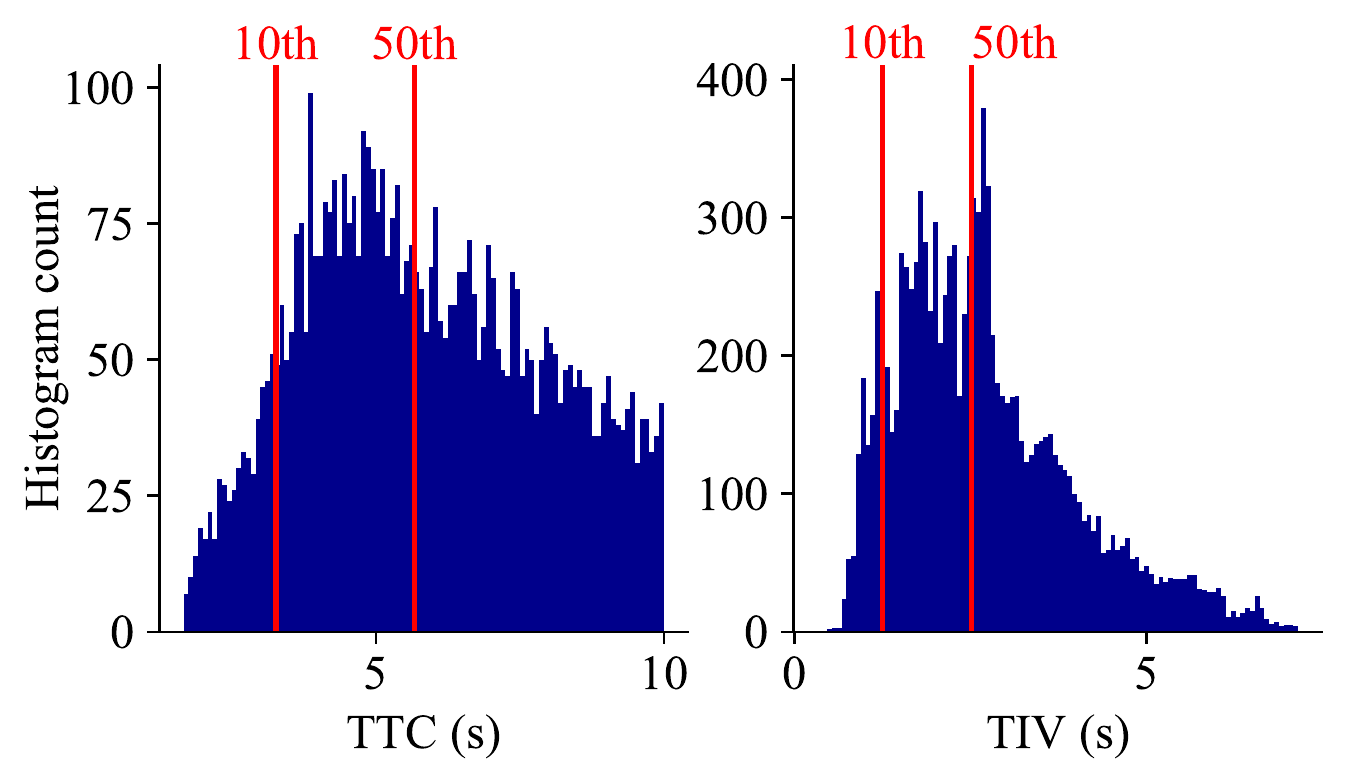}
    \caption{Left: occurrence frequency of TTC between the merging vehicles and their neighboring rear vehicle on the main road. Right: occurrence frequency of TIV. The red vertical lines mark the \textit{10th} and \textit{50th} percentile.}
    \label{fig:ttc_tiv_hist}
\end{figure}

\begin{figure}[t]
    \centering
    \includegraphics[width=0.9\linewidth]{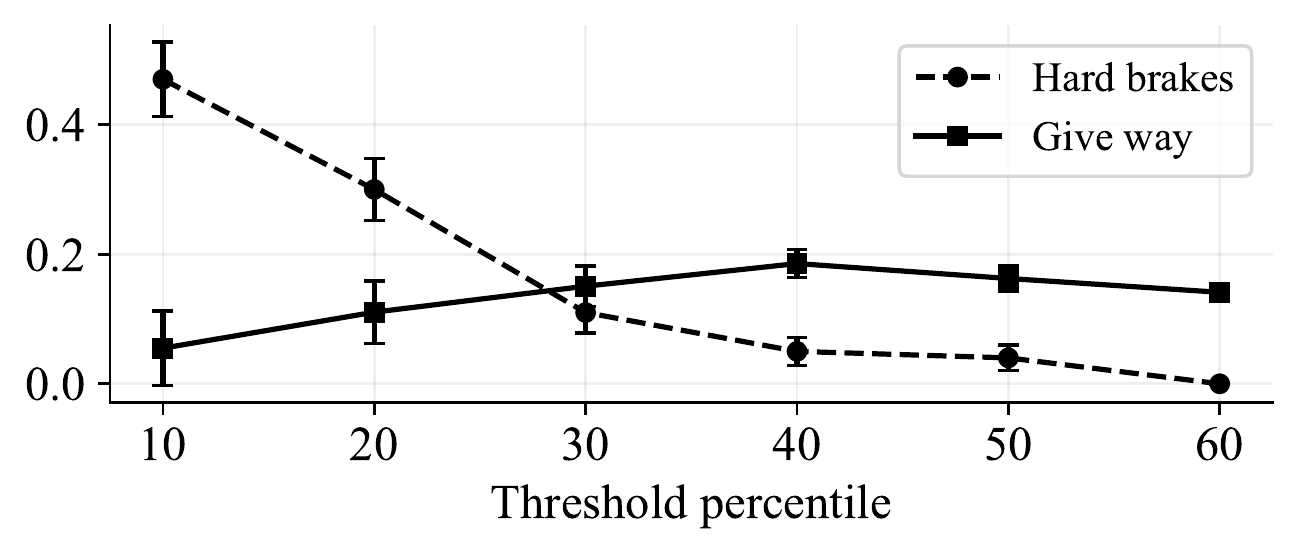}
    \caption{Rates of hard braking and ``give-way'' policies for the rule-based agent using different percentile ranks for the $\mathrm{TTC_{safe}}$ and $\mathrm{TIV_{safe}}$. The error bars indicate the standard error of the mean obtained by testing the agent on 100 episodes with randomly configured traffic environments, i.e., $\sigma_{\mathrm{sample}}/\sqrt{100}$, where $\sigma_{\mathrm{sample}}$ is the standard deviation for the 100 episodes.}\label{fig:param_testing}
\end{figure}

The thresholds $\{\mathrm{TTC_{safe}}, \mathrm{TIV_{safe}}, \mathrm{TTC_{min}}, \mathrm{TIV_{min}}\}$ have a strong influence on the agents' driving behavior. For example, assigning thresholds that are too high would result in the agents behaving conservatively, missing otherwise reasonable opportunities to initiate a merge. Conversely, small thresholds would result in agents expressing overly aggressive behaviors. 
Instead of tuning each threshold manually, we use the driving dataset $\mathcal{D}$ to determine reasonable values for our experiments. 
The occurrence frequencies of TIV and TTC for the merging vehicles in the dataset are shown in \cref{fig:ttc_tiv_hist}. The TTC values longer than \SI{10}{\second} were ignored as they can be considered too long to be of immediate concern for decision making. 
\cref{fig:param_testing} illustrates how, for the rule-based agent, the rate of hard braking actions of main-road vehicles and the rate at which the agent chooses to give way to other cars vary with the thresholds. We can see that almost half of the driving episodes result in a hard brake when the thresholds are low. With higher thresholds, the rule-based agent becomes more conservative, giving way to other drivers more frequently and causing fewer hard braking actions. For the rule-based agent, we use the \textit{50th} percentile in order to ensure safe driving without being overly conservative. For the other agents (which perform planning), we use the \textit{10th} percentile for the penalty thresholds $\mathrm{TTC_{min}}$ and $\mathrm{TIV_{min}}$ to discourage the agents from entering high-risk situations (see \cref{sec:reward_function} for details). The various parameters for the simulation environment and the configuration parameters for the agents are given in \cref{table:simulation_parameters}.

\section{Experiments and Results}

\subsection{Case Studies} \label{sec:qualitative}

\begin{figure}   
    \centering
    \includegraphics[width=\linewidth]{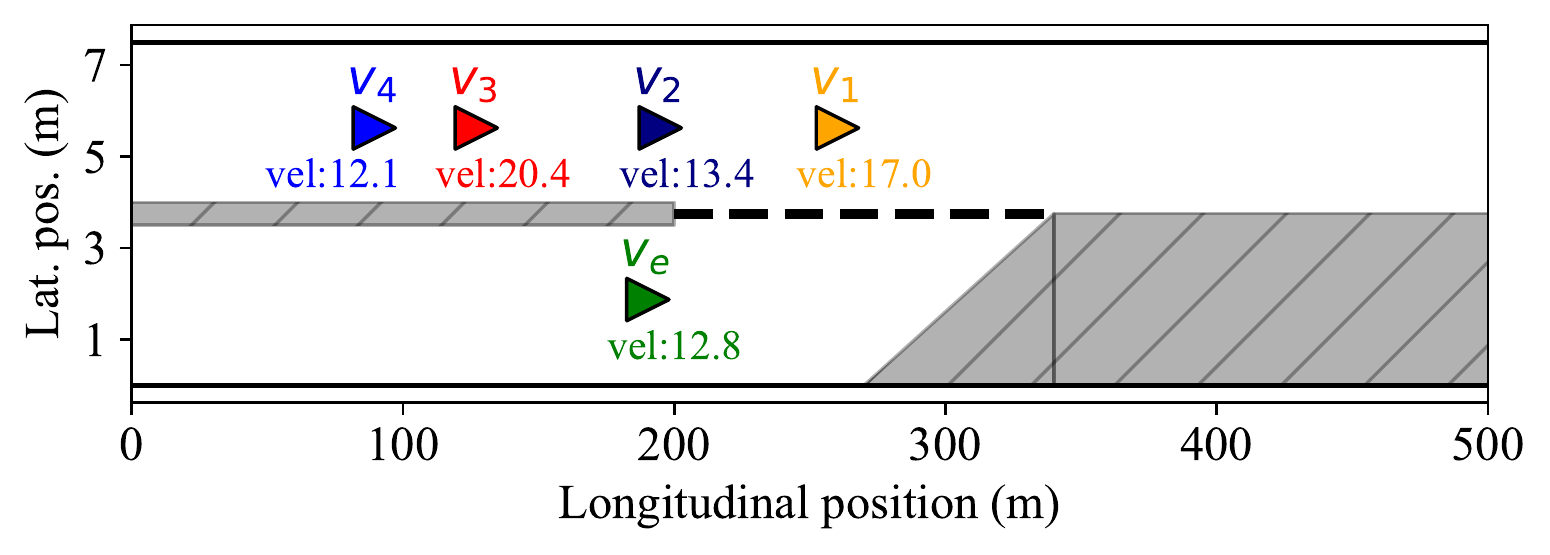}
    \caption{Initial state of the traffic environment with the agent (green) approaching the merging zone ahead of vehicle $v_3$.}\label{fig:scene_plot_initial}
\end{figure}

We introduce three driving scenarios and use them to test our planner, highlighting its ability to handle various traffic situations. The agent's goal in all the scenarios is to complete a merge maneuver in a timely manner while ensuring the safety of both itself and the other vehicles it encounters. The agent performs 150 planning iterations to evaluate its candidate policies at each decision step.\footnote{Our current Python implementation has an average runtime of approximately \SI{200}{\milli\second} per iteration on a \SI{3.2}{\giga\hertz} Intel i5-6500 CPU.} The test scenarios start with the initial state shown in \cref{fig:scene_plot_initial}, with the agent driving on the on-ramp at an initial speed of~\SI{12.8}{\meter\per\second}. The only difference between the three scenarios lies in the internal state of vehicle $v_3$, which is chosen such that the vehicle expresses either normal or aggressive driving behaviors.

\begin{figure} [t]
	\centering
	\subfloat[Scenario 1: encountering a normal driver.\label{fig:scene_plot_1}]{
		\includegraphics[trim={0.2cm 6.5cm 0.2cm 0.2cm}, clip=true, width=0.45\textwidth]{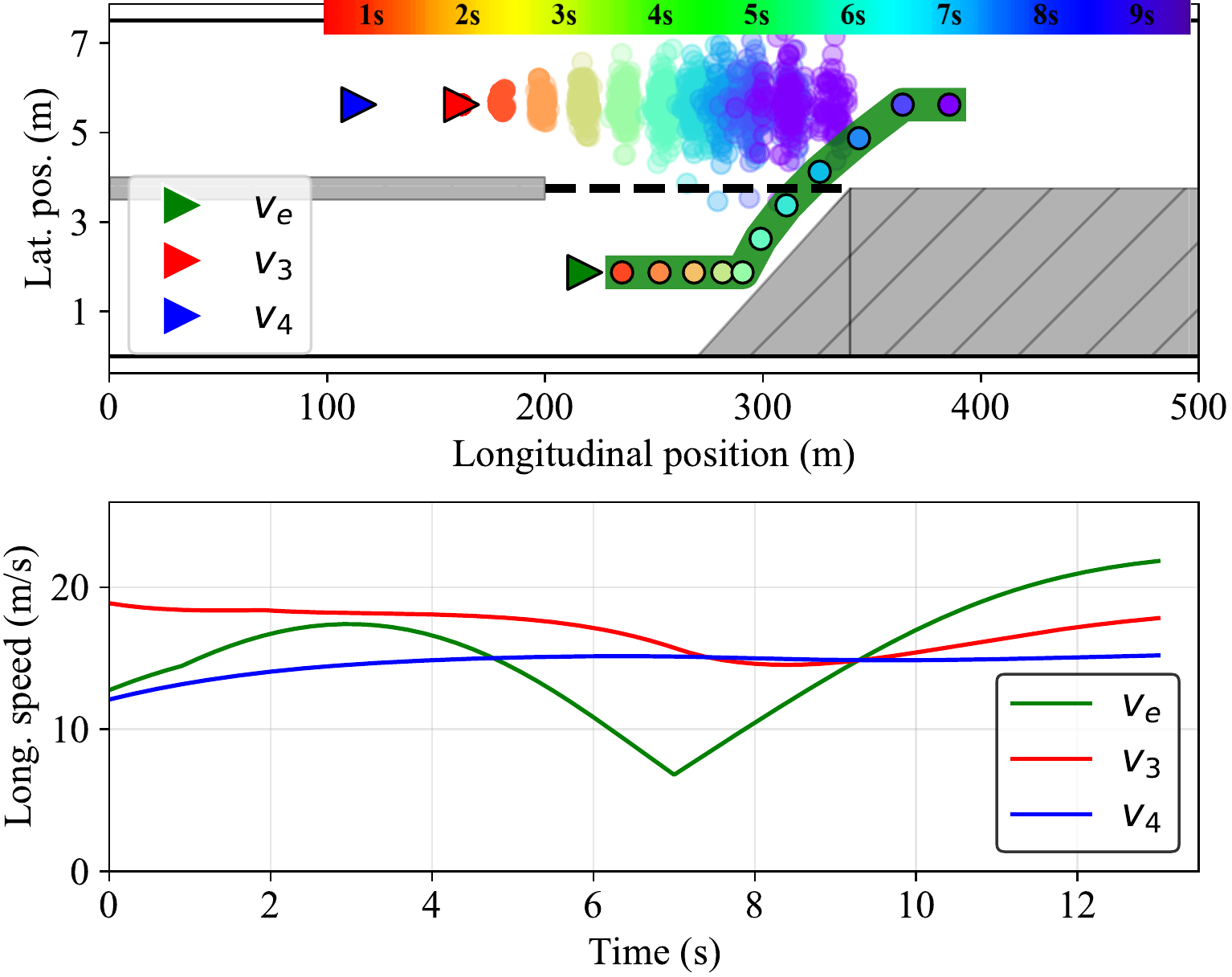}}   \\
	\subfloat[Scenario 2: encountering an aggressive driver. \label{fig:scene_plot_2}]{
		\includegraphics[trim={0.2cm 6.5cm 0.2cm 0.2cm}, clip=true, width=0.45\textwidth]{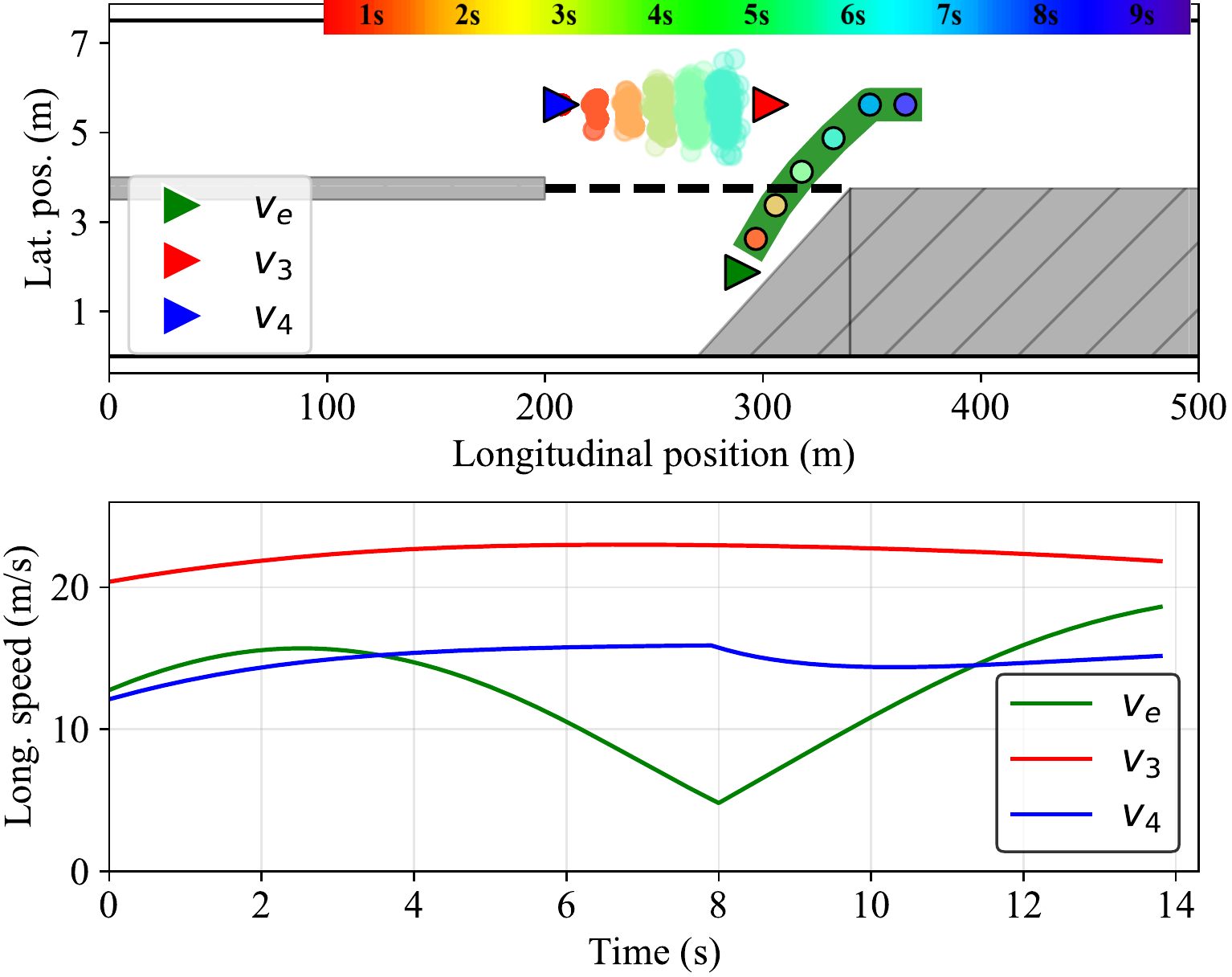}}
    \caption{Two example traffic scenarios, with the trajectories taken by the agent shown in green. The colored points in the top rows of the plots show the agent's anticipated future states of $v_3$. In (a), $v_3$ is a normal driver ($\psi_{v_3} = 0.5$). In (b), $v_3$ is an aggressive driver ($\psi_{v_3} = 0.9$). When dealing with a normal driver, the agent chooses to merge, having anticipated that $v_3$ is likely to yield. In the case of encountering an aggressive driver, the agent behaves defensively, waiting for $v_3$ to pass first before merging behind. Note that these driving strategies have not been pre-programmed but instead emerge from planning using our learned behavior model.}
	\label{fig:scene_plot}
\end{figure}

\subsubsection*{Scenario 1) encountering a normal driver}
 
\cref{fig:scene_plot_1} shows the simulation result for Scenario 1, where the agent (with label $v_e$) has to interact with a normal driver (shown in red). The trajectory shown in green results from the agent's actions, and the anticipated future states of $v_3$ are the colored points projected onto the road. 
The figure also plots the vehicles' speed profiles. As can be seen, the agent decides to merge into the available gap in front of $v_3$, having anticipated that $v_3$ is likely to yield. 
The agent initially chooses to speed up before slowing down when nearing the end of the merging zone. At about $t=\SI{7}{\second}$, the agent starts to move toward the main road with the ACC attempting to reach a desirable speed by continually adjusting the agent's longitudinal acceleration. The merge maneuver gets completed at $t=\SI{11.9}{\second}$. 
 
\subsubsection*{Scenario 2) encountering an aggressive driver}
 
In Scenario 2, the simulation episode starts with the same initial traffic state as Scenario 1, although $v_3$ is now an aggressive driver. The corresponding vehicle trajectories are plotted in \cref{fig:scene_plot_2}. At first, the agent chooses to accelerate. As the scene develops and the agent gets closer to the end of the merging zone, it decides to give way to $v_3$, having regarded a merge maneuver as unsafe.
After about $\SI{8}{\second}$ and once $v_3$ has passed, the agent initiates a merge and reaches the main road at $t=\SI{12.9}{\second}$.
 
 \begin{figure} [t]
	\centering
	\subfloat[Samples from the latent space in Scenario 1.\label{fig:latent_scenario_1}]{
		\includegraphics[trim={0.2cm 0.3cm 0cm 0.2cm}, clip=true, width=100pt]{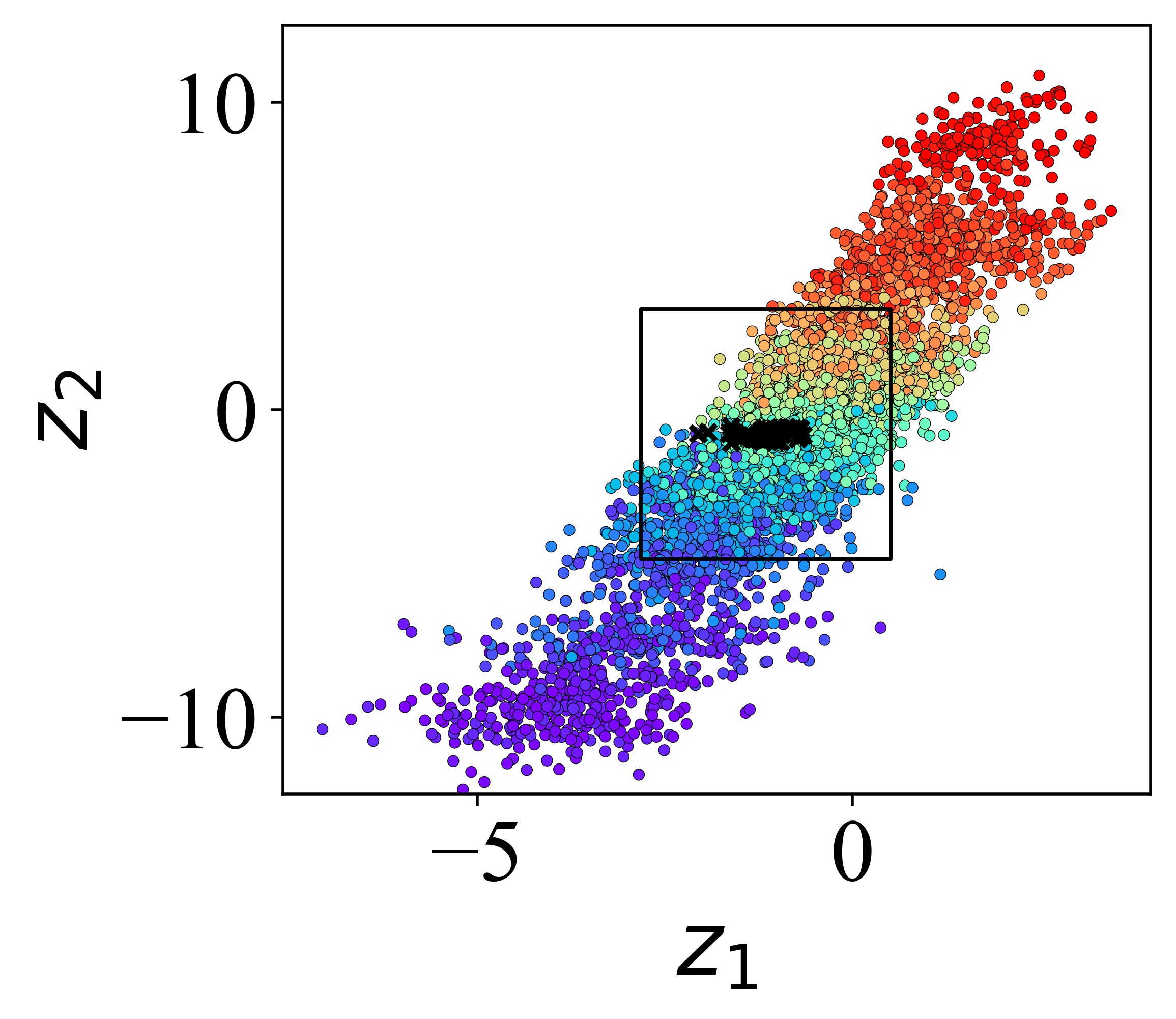}} \hspace{3pt}
	\subfloat[Samples from the latent space in Scenario 2.\label{fig:latent_scenario_2}]{
		\includegraphics[trim={0.2cm 0.3cm 0cm 0.2cm}, clip=true, width=128pt]{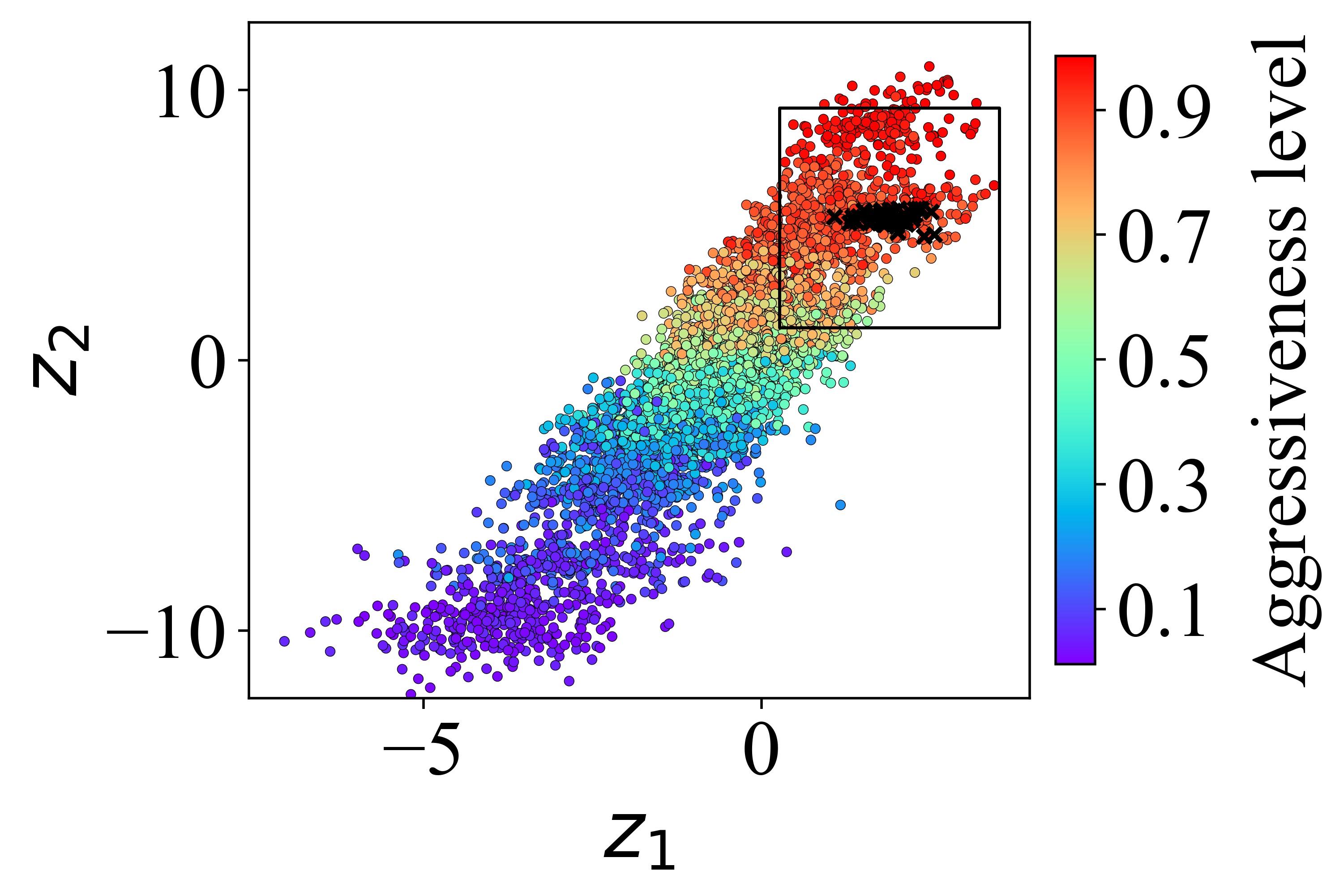}}  
    \caption{Visualization of the two-dimensional latent space with coloring according to the ground truth level of driver aggressiveness; the black points are samples from the latent space obtained for the respective scenarios.}  
	\label{fig:latent_scenarios_12}
\end{figure}

To gain insight into how the agent's belief about other drivers has influenced its decision to merge in Scenario 1 and give way in Scenario 2, we have visualized the model's latent space for both scenarios in \cref{fig:latent_scenarios_12}.  
The figure was produced using 5000 vehicle trajectories which we obtained using our traffic simulator. The points in the plot are colored according to driver aggressiveness, ranging
from purple (most timid) to red (most aggressive). 
The black points are the drawn samples from the latent space inferred specifically for vehicle $v_3$.
From inspecting the latent space, we can observe that the model has correctly identified the normal and aggressive driving tendency of $v_3$ in both scenarios, demonstrating its ability to form accurate beliefs about the behavior of other drivers. By using the model for planning, the agent is able to adapt its behavior to navigate both scenarios effectively.   
 
\begin{figure} [t]
    \centering
    \includegraphics[width=\linewidth, trim={0.2cm 0.3cm 0cm 0.2cm}, clip=true]{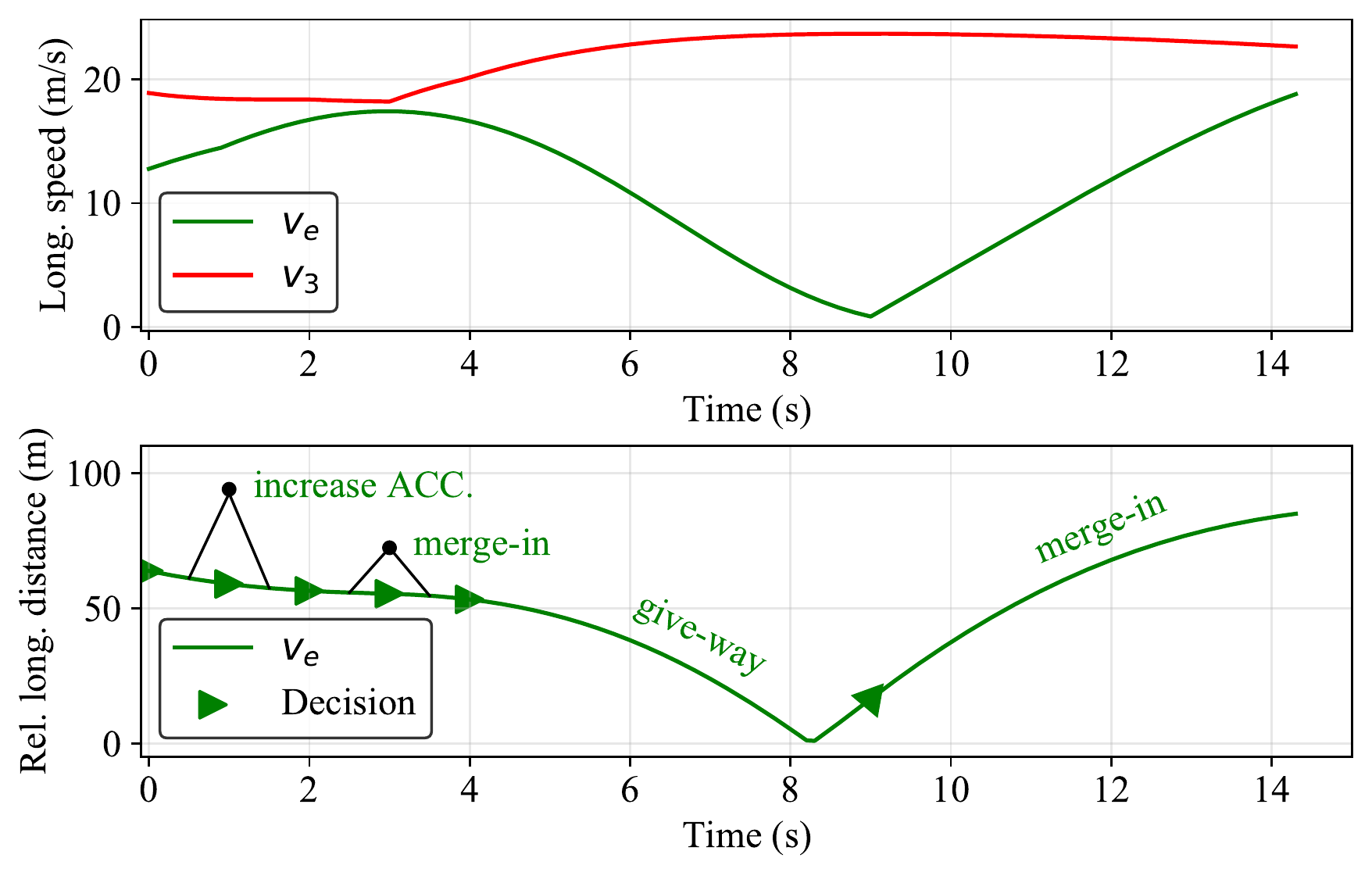}
    \caption{Planning results obtained for a merging scenario in which vehicle $v_3$ is an adversarial driver. At $t=\SI{3}{\second}$, $v_3$ becomes combative, accelerating to close the gap with its front neighbor. Bottom: longitudinal distance between the ego and $v_3$ with each decision labeled with the agent's chosen policy. Note that the number of time steps between two consecutive decision steps is not fixed and depends on each policy's termination condition.}\label{fig:adversarial}
\end{figure} 

\subsubsection*{Scenario 3) encountering an adversarial driver}

In Scenarios 1 and 2, the aggressiveness of drivers does not vary during the evaluation episode, meaning that the observed past behavior of drivers remains consistent with their future behavior. 
However, we can consider environments in which the behavioral pattern of some drivers is inconsistent across time. For instance, if autonomous cars gain a reputation for strictly prioritizing safety, it is reasonable to imagine that some drivers might become ``combative'' when encountering them, believing that autonomous cars are easier agents to deal with than other humans  \cite{tennant2017autonomous}.  
To mimic a similar adversarial setting, we initialize the environment with the same configuration as in Scenario 1, where $v_3$ is a normal driver, and then change the driver's internal state later in the episode. 
At $t=\SI{3}{\second}$, we turn $v_3$ into an aggressive driver. 

The planning results for Scenario 3 are shown in \cref{fig:adversarial}. 
At first, vehicle $v_3$ decelerates slightly, reaching a speed of about $\SI{18.5}{\meter\per\second}$ after two seconds. As in Scenario 1, the agent initially decides to accelerate, maintaining its distance with $v_3$. However, at $t=\SI{3}{\second}$, $v_3$ also begins to accelerate, narrowing the gap available for the agent to merge safely. As a result, the agent chooses to give way later in the episode, rapidly decelerating to a stop as it approaches the end of the merging zone. That is, when the agent finds that it cannot elicit cooperation from $v_3$, it changes its strategy by letting $v_3$ pass first. The agent can adapt its behavior under changing circumstances since each planning iteration starts with a belief state that contains information from the latest observations.

\subsection{Performance Results} \label{sec:quantitative}

We conduct quantitative evaluations to compare the performance of different decision making agents. Each agent is tested on 100 randomly generated traffic scenarios.
To expose an agent to diverse traffic conditions, we populate the road with $n$ vehicles where $n\in_R \{4, 5, 6, 7\}$ is the total number of vehicles on the road for a given episode.  
Note that while each evaluation episode is configured randomly, they are made identical for different agents to ensure a fair comparison. 
We use the following three metrics for evaluating performance: 
1) the rate at which an agent's actions lead to a safety violation; 2) the mean episode reward; and 3) the average time it takes for an agent to complete a merge maneuver.

\begin{figure} [t]
    \centering
    \includegraphics[width=0.9\linewidth]{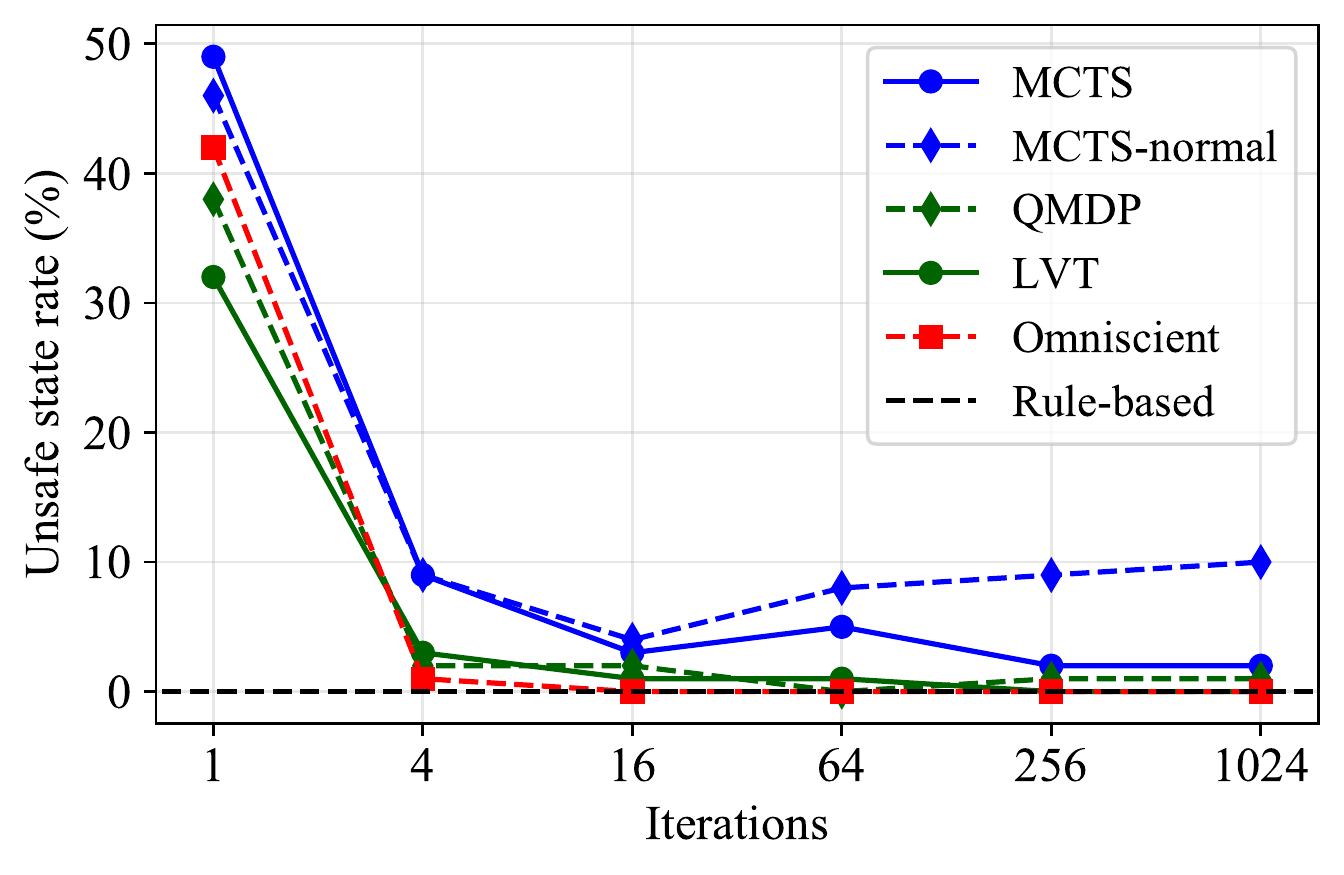}
    \caption{Rate of safety violations for a varying number of planning iterations.}\label{fig:planner_got_bad_state}
\end{figure} 

\begin{figure}   
    \centering
    \includegraphics[width=0.9\linewidth]{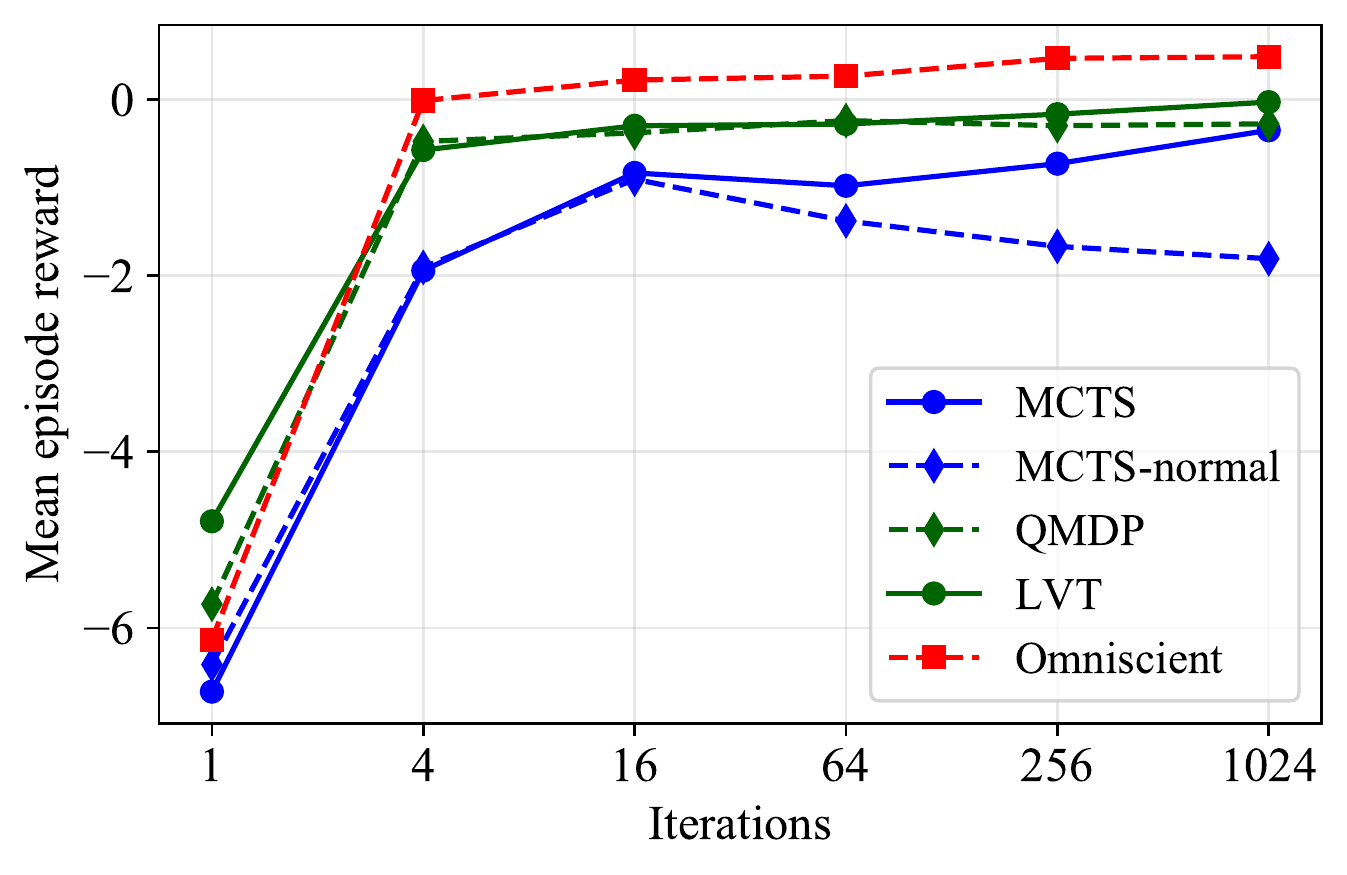}
    \caption{Mean episode reward received by agents using different planning algorithms as a function of the number of planning iterations. The error bars are omitted for clarity, but the margins are relatively small.}\label{fig:planner_rewards}
\end{figure} 

\begin{figure}    
    \centering
    \includegraphics[width=0.9\linewidth]{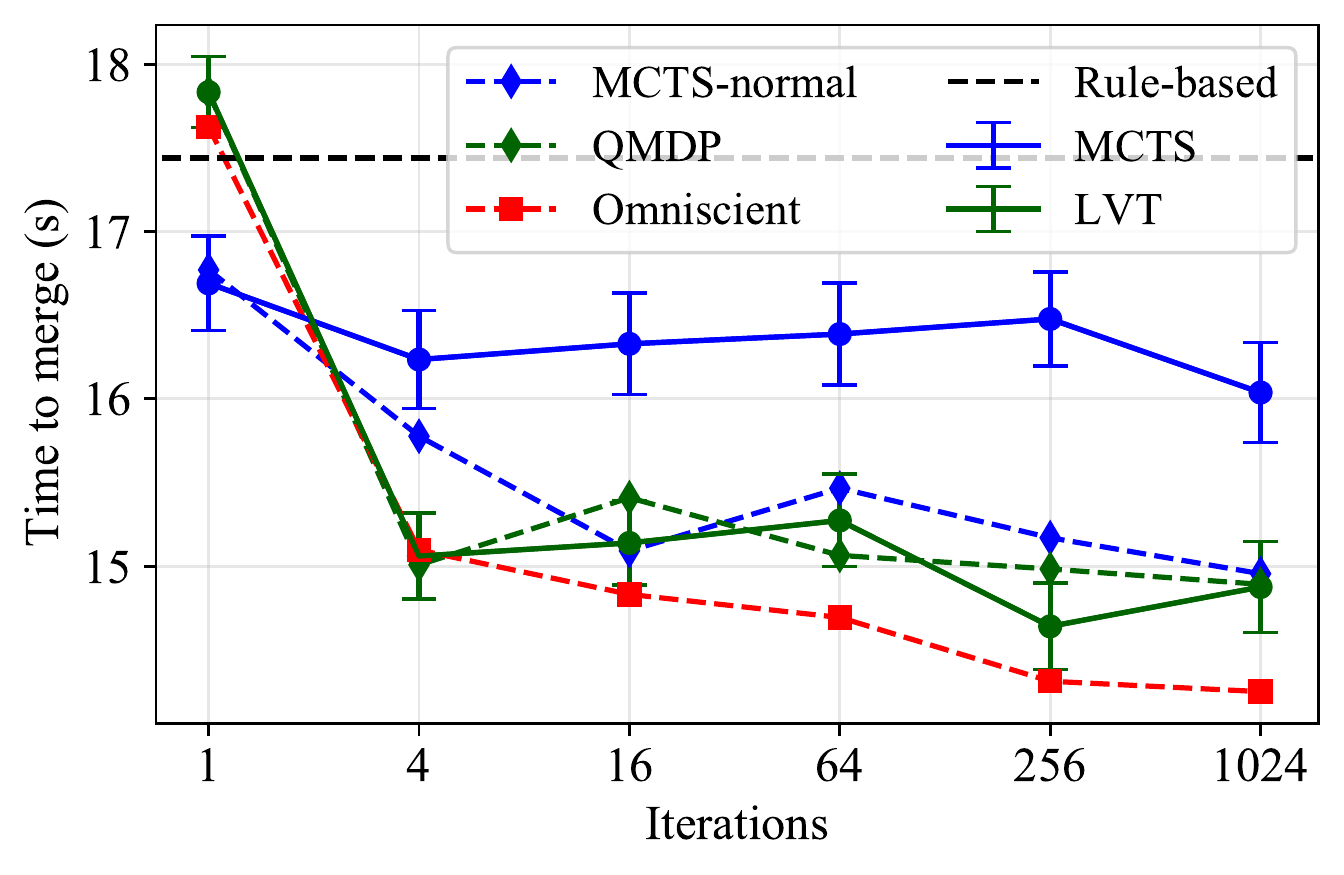}
    \caption{Average time it took different agents to complete their merge maneuvers. For the LVT and MCTS agents, the error bars indicate the standard error of the mean. The error bars for the other agents are omitted for improve clarity, but the margins are relatively small.}\label{fig:planner_ttm}
\end{figure}

\cref{fig:planner_got_bad_state} illustrates the effect number of planning iterations can have on the safety of the agents' driving strategies. 
The rule-based agent never enters an unsafe state, which is expected as it is designed to express a safe driving behavior (see \cref{sec:baseline_methods}). 
For all other agents except for MCTS-normal, 
fewer than $5 \%$ of episodes result in a safety violation after only 16 iterations. The LVT agent, in particular, reaches a safety violation rate of zero. The risky behavior of the MCTS-normal agent can be explained by its biased and over-confident expectation that all drivers on the road follow a normal driving style; this becomes fatal when an approaching vehicle on the main road, whose behavior ultimately determines whether or not the agent should merge, drives aggressively. 

\cref{fig:planner_rewards} plots the average reward accumulated using different planning algorithms as a function of the number of planning iterations. With only one iteration, which amounts to choosing policies at random, the agents receive high penalties. With a few iterations, all agents attain a significant performance improvement. 
The LVT and QMDP produce comparable results, although the performance of LVT improves further with additional planning iterations while performance settles for QMDP after 64 iterations. This result suggests that, on average, taking information-gathering actions does not offer a significant advantage in the merging scenario considered.  
From the plot, we can also see that the reward accumulated by MCTS-normal starts to decrease after 16 iterations. This result is surprising, as we expect additional iterations to improve performance. 
The fact that the MCTS-normal agent chooses its actions assuming all other drivers are \textit{normal} may explain this phenomenon; the agent is overly confident and does not hedge against the diversity of driver behaviors.  
 
\cref{fig:planner_ttm} shows the average time it took different agents to complete their merge maneuvers. Here we have only considered the episodes that did not lead to a safety violation. 
The rule-based agent takes the longest time, indicating that it follows an overly conservative strategy in many traffic instances.  
By performing more assertive merges, the LVT agent is able to close the performance gap to the Omniscient agent, with the QMDP agent showing similar results. The MCTS agent takes longer to complete its merge maneuvers and remains defensive even when performing additional planning iterations. This trend can be explained by our chosen reward function, which heavily penalizes dangerous and uncooperative agent actions; in the face of uncertainty and unable to learn about other drivers, the MCTS agent opts for an overly conservative driving strategy.  

\section{Conclusions}

In this paper, we proposed a decision making approach for autonomous driving, focusing on a ramp merging case study where it is important to consider the uncertain interactions among vehicles. We framed the problem as a POMDP and derived locally optimal actions for the autonomous vehicle via online planning. Our method leverages models learned from driving data to hold a belief about the uncertain behavior of other drivers and predict their future motion.
We show that the planning algorithm generates safe driving behaviors and outperforms several baselines. 
Moreover, simulation experiments reveal that the approach is robust enough to handle various situations, even when dealing with an adversarial driver who becomes combative when interacting with the autonomous vehicle; it results, for example, in the autonomous vehicle postponing its decision to merge, having anticipated a probable unsafe situation to arise in the future. 
This ability to navigate challenging encounters with other drivers is crucial for the large-scale integration of autonomous vehicles on public roads. 

Future work should focus on improving the computational efficiency of the algorithm, for example, by applying appropriate function approximation techniques to reduce the search depth for planning \cite{hoel2019combining} and lowering the computational cost of the model. Further, the approach can be extended to consider not only the uncertain behavior of nearby drivers but also to generalize to environments with other traffic objects, including parked cars and pedestrians.
  

\bibliography{my_refs}


\phantomsection
\vskip -2\baselineskip plus -1fil
\begin{IEEEbiography}[{\includegraphics[trim={3.5cm 4cm 3.5cm 1cm}, clip=true, clip, width=1in,height=1.25in, keepaspectratio]{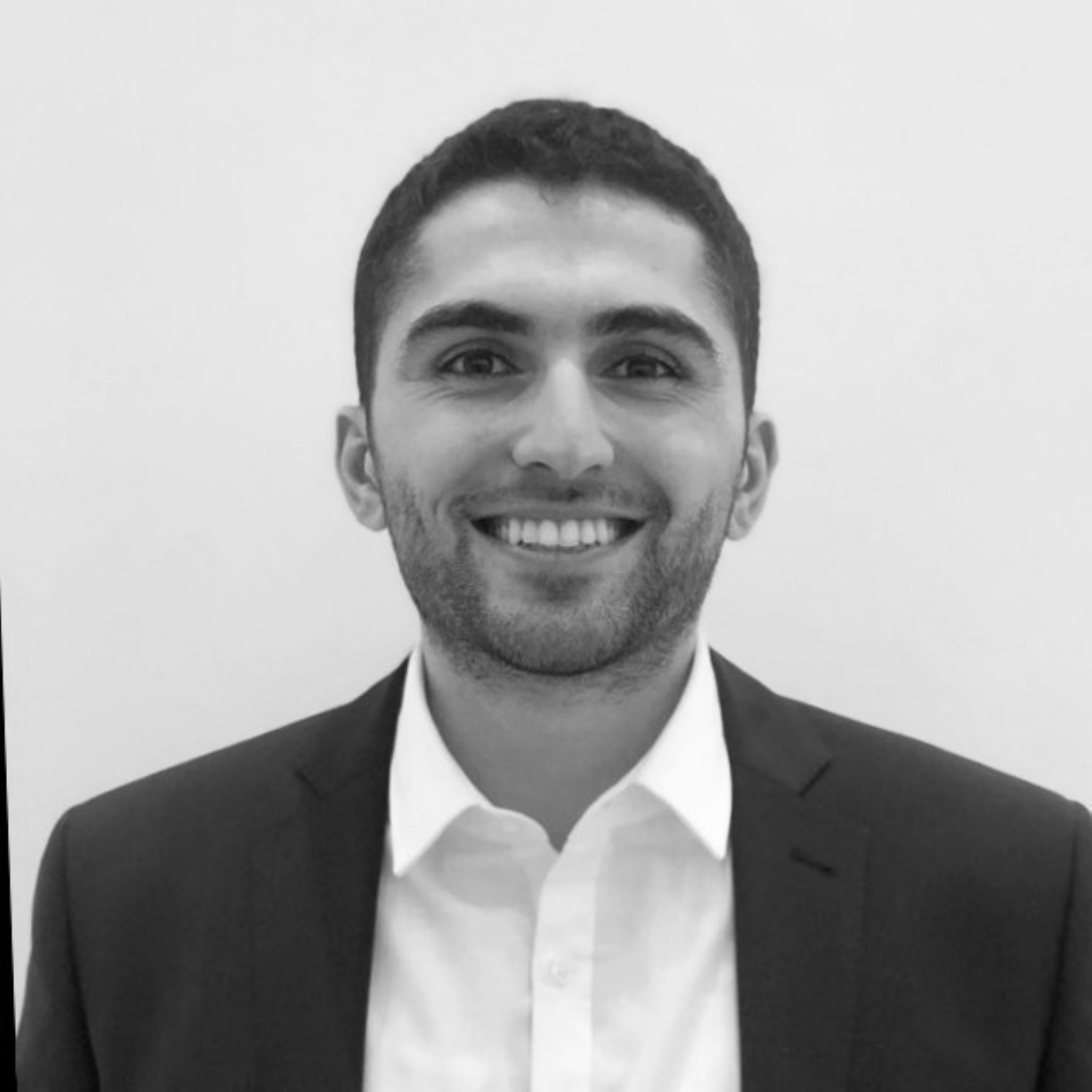}}]{Salar Arbabi}
received the M.Eng. degree in Mechanical Engineering from the University of Surrey, U.K., in 2018. He is currently pursuing the Ph.D. degree with the Centre for Automotive Engineering, University of Surrey. His research focuses on decision making and motion planning under uncertainty for autonomous driving. 
\end{IEEEbiography}

\vspace{7cm}
\begin{IEEEbiography}[{\includegraphics[width=1in,height=1.25in,clip, keepaspectratio]{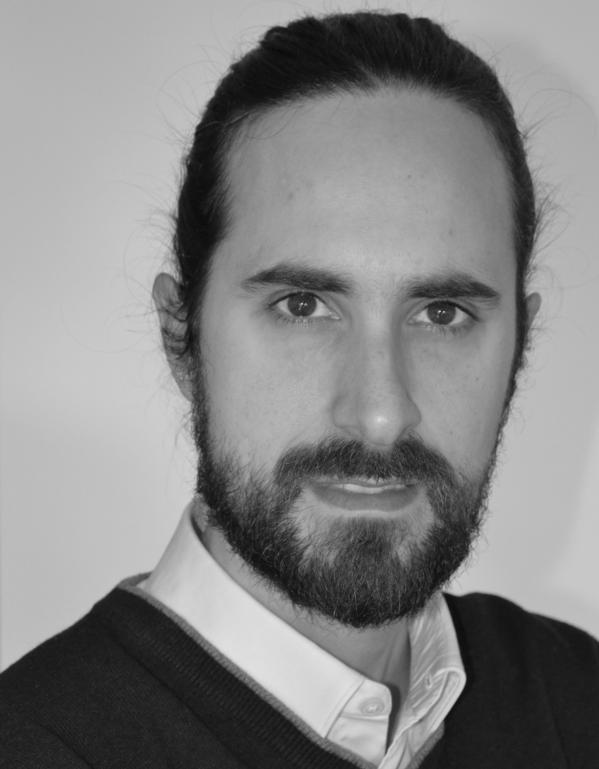}}]{Davide Tavernini} received the M.Sc. degree in mechanical engineering and Ph.D. degree in dynamics and design of mechanical systems from the University of Padova, Padua, Italy, in 2010 and 2014. During his Ph.D. he was part of the motorcycle dynamics research group.  He is a Senior lecturer in advanced vehicle engineering with the University of Surrey. His research interests include vehicle dynamics modeling, control and state estimation, mostly applied to electric vehicles with multiple~motors.
\end{IEEEbiography}

\vskip -2\baselineskip plus -1fil
\begin{IEEEbiography}[{\includegraphics[width=1in,height=1.25in,clip, keepaspectratio]{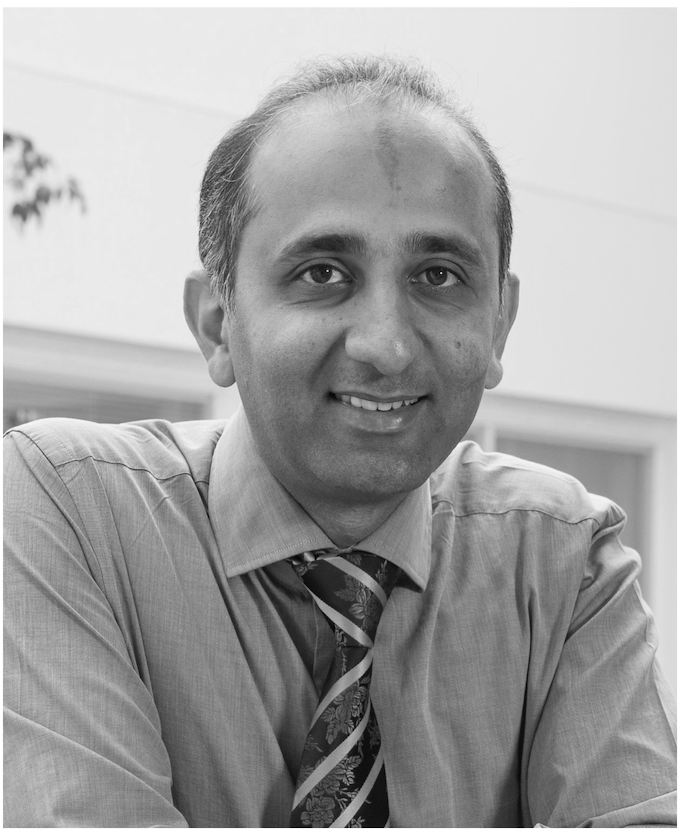}}]{Saber Fallah} leads the Connected and Autonomous Vehicles Laboratory (CAV-Lab) at the University of Surrey, where his team conducts cutting-edge research on developing trustworthy autonomous robotics and vehicles. The focus of his research is on developing safe and explainable AI and autonomy. He has secured research funding from EPSRC, Innovate UK, EU, KTP, and industry. His work has contributed to the state-of-the-art research in the areas of autonomous decision making and learning optimal control for autonomous systems.
\end{IEEEbiography}

\vskip -2\baselineskip plus -1fil
\begin{IEEEbiography}[{\includegraphics[trim={1cm 1cm 1cm 0cm}, clip=true, clip, width=1in,height=1.25in, keepaspectratio]{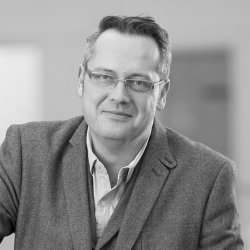}}]{Richard Bowden} (Senior Member, IEEE) is Professor of computer vision and machine learning with the University of Surrey, where he leads the Cognitive Vision Group within the Centre for Vision, Speech and Signal Processing. His research centers on the use of computer vision to locate, track, and understand humans. He is a fellow of the Higher Education Academy and the International Association of Pattern Recognition (IAPR). He has previously held a Royal Society Leverhulme Trust Senior Research Fellowship and served as Associate Editor for the Journals Image and Vision computing and IEEE Transactions on Pattern Analysis and Machine Intelligence.
\end{IEEEbiography}

\bibliographystyle{IEEEtran}
\end{document}